\documentclass[times,twocolumn,final]{elsarticle}

\usepackage{medima}
\usepackage{framed,multirow}
\graphicspath{{figures/}}

\usepackage{amssymb}
\usepackage{latexsym}
\usepackage{pifont}
\newcommand{\xmark}{\ding{55}}
\newcommand{\cmark}{\ding{51}}

\usepackage{url}
\usepackage{xcolor}
\usepackage{subfigure}
\usepackage{soul}

\usepackage{booktabs} 
\usepackage{amsmath,url,graphicx,amssymb}
\usepackage{amsfonts}
\usepackage{multirow}
\usepackage{algpseudocode}
\usepackage{pifont}
\usepackage{color}
\usepackage{xspace}
\usepackage{cite}
\usepackage{colortbl}
\usepackage{xcolor}
\usepackage{bm}
\usepackage{bbm}
\usepackage{todonotes}
\usepackage[
  separate-uncertainty = true,
  multi-part-units = repeat
]{siunitx}

\newcommand{\eg}{\textit{e.g.}}

\newcommand{\ie}{\textit{i.e.}}

\newcommand{\mR}{\mathcal{R}}
\newcommand{\mE}{\mathbb{E}}
\newcommand{\mX}{\mathcal{X}}
\newcommand{\mL}{\mathcal{L}}
\newcommand{\mS}{\mathcal{S}}
\newcommand{\mT}{\mathcal{T}}

\definecolor{sblue}{HTML}{02BCD4}
\definecolor{sred}{HTML}{F44436}
\definecolor{spink}{HTML}{E91E62}
\definecolor{sgreen}{HTML}{8BC34A}
\definecolor{spurple}{HTML}{3F51B5}
\definecolor{slightgreen}{HTML}{CCDE3A}
\definecolor{sorange}{HTML}{FE9800}
\definecolor{sgolden}{HTML}{FFC108}

\usepackage{amsmath}
\usepackage{amssymb}
\usepackage{dsfont}
\newcommand{\ours}{\texttt{CCSI}\xspace}
\newcommand{\xl}[1]{{\color{red}[XL: #1]}}

\usepackage{hyperref}

\definecolor{newcolor}{rgb}{.8,.349,.1}

\journal{Medical Image Analysis}

\begin{document}

\verso{Sana Ayromlou \textit{et~al.}}

\begin{frontmatter}

\title{CCSI: Continual Class-Specific Impression for Data-free Class Incremental Learning}

\author[1,3]{Sana \snm{Ayromlou}\fnref{fn1}}
\fntext[fn1]{The work was completed at the University of British Columbia.}
\ead{s.ayromlou@ece.ubc.ca}

\author[2]{Teresa \snm{Tsang}} \ead{t.tsang@ubc.ca}

\author[1]{Purang \snm{Abolmaesumi}} \ead{purang@ece.ubc.ca}

\author[1,3]{Xiaoxiao \snm{Li}\corref{cor1}}
\ead{xiaoxiao.li@ece.ubc.ca}
\cortext[cor1]{Corresponding author: Xiaoxiao Li}

\address[1]{Electrical and Computer Engineering Department, The University of British Columbia, Vancouver, BC V6T 1Z4, Canada}
\address[2]{Vancouver General Hospital, Vancouver, BC V5Z 1M9, Canada}
\address[3]{Vector Institute, Toronto, ON M5G 0C6, Canada}

\received{3 Aug 2023}
\finalform{2 Jun 2024}
\accepted{6 Jun 2024}

\begin{abstract}
In real-world clinical settings, traditional deep learning-based classification methods struggle with diagnosing newly introduced disease types because they require samples from all disease classes for offline training. Class incremental learning offers a promising solution by adapting a deep network trained on specific disease classes to handle new diseases. However, catastrophic forgetting occurs, decreasing the performance of earlier classes when adapting the model to new data. 
Prior proposed methodologies to overcome this require perpetual storage of previous samples,  posing potential practical concerns regarding privacy and storage regulations in healthcare. To this end, we propose a novel data-free class incremental learning framework that utilizes data synthesis on learned classes instead of data storage from previous classes.
Our key contributions include acquiring synthetic data known as \textit{Continual Class-Specific Impression} (\ours) for previously inaccessible trained classes and presenting a methodology to effectively utilize this data for updating networks when introducing new classes. 
We obtain \ours by employing data inversion over gradients of the trained classification model on previous classes starting from the mean image of each class inspired by common landmarks shared among medical images and utilizing continual normalization layers statistics as a regularizer in this pixel-wise optimization process.
Subsequently, we update the network by combining the synthesized data with new class data and incorporate several losses, including an intra-domain contrastive loss to generalize the deep network trained on the synthesized data to real data, a margin loss to increase separation among previous classes and new ones, and a cosine-normalized cross-entropy loss to alleviate the adverse effects of imbalanced distributions in training data.
Extensive experiments show that the proposed framework achieves state-of-the-art performance on four of the public MedMNIST datasets and in-house echocardiography cine series, with an improvement in classification accuracy of up to 51\% compared to baseline data-free methods. Our code is available at  \href{https://github.com/ubc-tea/Continual-Impression-CCSI}{https://github.com/ubc-tea/Continual-Impression-CCSI}.
\end{abstract}


\end{frontmatter}



\section{Introduction}

Current deep-learning models for medical imaging classification tasks have shown promising performance. Most of these models necessitate gathering all the training data and specifying all of the classes before the training. 
They train a deep learning model once during deployment and anticipate it to execute on all subsequent data.
However,  such requirement limits the feasibility in practical clinical settings where medical image data are continuously collected and change over time, \eg, when new disease types appear.

A promising approach to address this challenge in machine learning is to enable the system to engage in continual or lifelong learning, meaning that the deployed model can adapt to new data while maintaining the information gained from previously seen data.
Incorporating these learning techniques will make deep learning models more flexible to the constant expansion of medical datasets.
Medical continual learning has been utilized in various incremental scenarios~\citep{van2022three} by taking into account the non-stationary nature of incoming data. These scenarios include task incremental learning~\citep{gonzalez2023lifelong,liao2022muscle,xu2022expert,kaustaban2022characterizing,chakraborti2021contrastive}, where new medical tasks are introduced, \eg, extending a segmentation network to another body region; class incremental learning~\citep{chee2023leveraging,yang2021continual,li2020continual}, where new classes are added to the model, \eg, introducing a new disease type in the classification task; and domain incremental learning~\citep{yang2023few,srivastava2021continual,bayasi2021culprit}, where the model is faced with new medical domains that it has not been trained on. The majority of these incremental learning scenarios presume access to all the training data for the prior model or a part of it, which is stored in the memory systems. The re-training process then occurs on the saved data along with new data. 

\begin{figure}[t]
	\centering
	\includegraphics[width=0.4\textwidth]{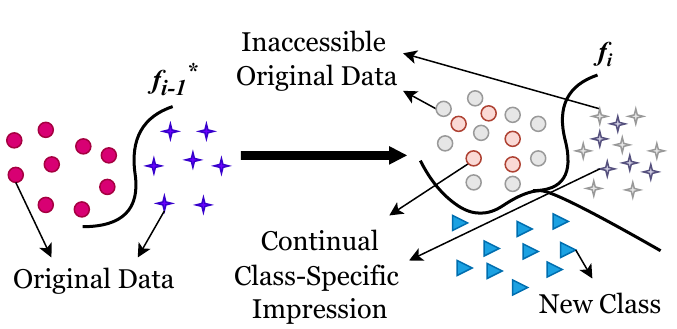}
	\caption{Representation of data-free class incremental learning. $f^*_{i-1}$ is the model trained on previous data, while $f_{i}$ is the updated model with new classes. This approach enables the incremental learning of new classes added to a previously trained model without having access to previous data. We propose to tackle this problem by synthesizing samples of previous classes as the continual class-specific impression and adding them to the continual training paradigm. Best viewed in coloured print. 
}
	\label{abstract_image}
\end{figure}

This research emphasizes class incremental learning in medical image analysis due to the growing fascination with deep learning-driven disease diagnosis applications and the distinct difficulty of expanding the label space. In the clinical deployment of medical image analysis, there are significant challenges posed by feasibility constraints related to the storage of medical data and strict privacy regulations like the Health Insurance Portability and Accountability Act (HIPAA) and the General Data Protection Regulation (GDPR)~\citep{voigt2017eu, o2004health}. These constraints often prohibit the application of many existing continual learning strategies that rely on accessing previous training data. As a result, in the medical domain, traditional continual learning approaches may not be allowed or feasible, making it essential to develop new strategies that can effectively handle data-free scenarios, which involves the introduction of new classes over time without having access to data from previous classes. The aim of data-free class incremental learning is to achieve optimal performance across both the original classes and the new class without accessing the original data for continuous training (depicted in Fig.~\ref{abstract_image}). 
To this end, researchers have put forth the concept of synthetic data as a viable alternative for data transformation and storage in natural image analysis~\citep{choi2021dual,pourkeshavarzi2021looking,smith2021always}. We have observed that directly applying these existing data synthesis methods for data-free class incremental learning on medical images failed to achieve the desired results (see Sec.~\ref{sec:baseline}) and recognized the challenges came from:
1) complex and high-dimensional nature of the medical data, and
2) intricate and often indistinguishable patterns among different classes, and
3) imbalanced class distributions and feature shifts between synthesized and real images, which makes it more prone to catastrophic forgetting.
Motivated by these observations and data synthesis challenges, we introduce a novel data-free class-incremental learning 
\textcolor{black}{pipeline} for medical images that has two main steps for learning newly introduced classes: \textcolor{black}{1) synthesizing data points for non-accessible previous classes by solving an inverse problem over the existing trained model which incorporates statistics of normalization layers and class-specific medical image priors, and 2) training on both synthetic data and new samples by introducing multiple novel loss functions to alleviate catastrophic forgetting and enhance knowledge transfer among tasks.} 
 \textcolor{black}{Specifically, inspired by mentioned unique characteristic of medical images, first, we propose to initialize synthetic images generation with the mean of each class as meta-data. Next, we regularize the pixel-wise optimization by matching the statistics in the normalization layer.
Furthermore, we propose leveraging a unique normalization layer named Continual Normalization (CN) in our data synthesis process to restore the \textit{Continual Class-Specific Impression (\ours)} from the previous classes. This proposition stems from our findings that indicate the critical role of high-quality synthetic images in Step 1 for maintaining prior task knowledge. This reliance is heavily based on the statistics stored in the respective normalization layers employed within the deep network. CN mitigates the impact of overwriting the statistics of the normalization layer for samples from newly introduced classes, thereby alleviating catastrophic forgetting of previous classes. Finally, we enhance the utility of this synthetic data in Step 2, by introducing a) \textit{intra-domain contrastive loss}, a semi-supervised domain adaptation approach, which effectively addresses the distribution mismatch between real and synthetic data, b) \textit{margin loss} which encourages robust decision boundaries between previous and new classes, and c) \textit{cosine-normalized cross-entropy loss} to handle imbalanced quality issue, leading to catastrophic forgetting in the data-free class incremental scenario.} Our contributions are summarized as follows:

\begin{itemize}
\item We propose \ours{}, a two-step pipeline of data-free class incremental learning for medical image analysis, featured by our novel data restoring strategy and loss functions. Our method fills the gap in this underexplored field of medical imaging \textcolor{black}{by leveraging medical image priors as meta-data to initialize our data synthesis step.}

\item We formulate data restoring as an inverse problem from model weights and leverage CN statistics in our pipeline \textcolor{black}{for the first time in the data-free class-incremental learning} to simultaneously improve data synthesis with CN parameter priors and alleviate overwriting parameters in continual model training.

\item \textcolor{black}{During data-free continual training, to enhance the utility of synthetic data, we introduce an intra-domain contrastive loss to overcome domain shift between synthesized and original images. Additionally, we utilize margin loss and cosine-normalized cross-entropy loss to reinforce decision boundaries and balance class significance.}

\item We conduct extensive experiments on four public datasets and one in-house dataset, demonstrating the effectiveness of \ours{} over the state-of-the-art baselines originally designed for natural image classification. 
\end{itemize}


Recently, we presented a preliminary version of this work based on \textsc{Class Impression}~\citep{ayromlou2022class}, where we proposed a similar two-step approach by adopting Batch Normalization (BN) layers' statistics instead of CN layers. This manuscript extends the preliminary work in two ways. First, we improve both steps in our preliminary version by introducing CN for better synthetic image generation and catastrophic forgetting mitigation. We note that the mean and variance computed in the BN layers are overwritten in Step 2, leading to the loss of valuable information. Therefore, the newly utilized normalization technique helps maintain class-specific information of continually introduced classes. Second, we extend extensive experiments on four new datasets and detailed qualitative and quantitative ablation studies to validate the effectiveness of $\ours$.     

\section{Related Works}
\label{related}

\subsection{Strategies in Continual Learning}

The objective of continual learning in deep learning is to develop techniques that enable deep networks to effectively learn new tasks and data while retaining knowledge from previously learned tasks~\citep{masana2020class}. However, when a deep learning model is trained on new data, it frequently suffers from a phenomenon called catastrophic forgetting, where it tends to forget previously acquired knowledge ~\citep{article}. To address this problem, recent efforts have concentrated on four distinct approaches:

\noindent \emph{1) Rehearsal-based methods}, which alleviate forgetting  by revisiting stored samples from prior tasks~\citep{cha2023rebalancing,pham2022continual,bang2021rainbow,rolnick2019experience,isele2018selective,rebuffi2017icarl} or training generative network to generate samples~\citep{ramapuram2020lifelong,lavda2018continual,shin2017continual}. 

\noindent \emph {2) Regularization-based methods}, which use defined regularization terms such as additional losses to preserve previously learned knowledge while updating on new tasks~\citep{zhang2020class,hou2018lifelong,li2017learning,rannen2017encoder,jung2016less} or assign an importance value to weights of the network and modify them accordingly~\citep{titsias2019functional,schwarz2018progress,kirkpatrick2017overcoming,zenke2017continual};

\noindent \emph{3) Bias-correction methods}, which aim to overcome forgetting by reducing task recency bias by calibrating classification~\citep{ahn2021ss,wu2019large,castro2018end}; 

\noindent \emph{4) Parameter isolation methods}, which allocate a fixed part of a static architecture for each task and only update that part during training~\citep{yoon2019scalable,serra2018overcoming,yoon2017lifelong,aljundi2017expert,rusu2016progressive}.

It is important to note that some works incorporate elements from multiple categories. For example, \citet{buzzega2020dark,douillard2020podnet} and \citet{castro2018end} used a combination of rehearsal methods and regularization techniques, while \citet{hou2019learning} also combined some bias correction techniques over them to overcome catastrophic forgetting. Among these approaches, the rehearsal-based strategies have been shown to produce the best results~\citep{hou2019learning}. However, in order to perform re-training on the saved data alongside the new data, they require accessing either the entire dataset or a portion of the data used to train the previous model.

\noindent\textcolor{black}{While some methods continue to train the same static model on new data~\citep{buzzega2020dark}, others propose creating new modules dynamically to handle the growing training distribution~\citep{yan2021dynamically}. More specifically, they train a new model each time a new class is introduced. One main drawback of these methods is that continually adding new modules causes unaffordable memory overhead. Additionally, directly retaining old modules leads to noise in the representations of new categories, harming performance in new classes~\citep{wang2022foster}. Some recent works have introduced various compression methods~\citep{wang2022foster,wang2022beef} or have suggested adding only later modules of the network~\citep{zhou2022model}. As we believe such a network expansion can be orthogonal to our proposed data-free approach, we only focus on static architectures in this work due to the complexity of dynamic networks.}
\subsection{Data-free Continual Learning}
Recently, there have been proposed approaches in the field of \textit{prototype-based} and \textit{data-free} methods to achieve high performance without the need to store large amounts of data. However, prototype-based approaches typically assume access to a large dataset from the first task for pre-training the model~\citep{sun2023exemplar,zhu2021prototype,yang2021continual,wu2021striking,yu2020semantic}. Consequently, these approaches may fail when such pre-trained models are not available.
While some data-free approaches, which learn prompts within frozen and pre-trained transformers, have shown effectiveness in natural image datasets, their application to medical images may be challenging due to the lack of access to a pre-trained encoder that can effectively separate the data~\citep{smith2023coda,wang2022learning,wang2022dualprompt}. There has been another category of data-free approaches introduced, which generate images for previous classes using Deep Inversion without the need for any prior pre-trained model~\citep{ayromlou2022class,choi2021dual,smith2021always}. In this study, we additionally emphasize the utilization of a Deep Inversion to retrieve samples from previous classes, as it does not require the assumption of having access to a pre-trained model on medical data.

\subsection{Medical Image Synthesize}

High-quality healthcare data holds immense importance for conducting high-quality research, driving better development initiatives, achieving favourable outcomes, facilitating well-informed medical decisions, and ultimately improving the overall quality of life~\citep{murtaza2023synthetic}.  Synthetic medical data has progressively replaced the original data in various applications~\citep{wang2021review}, encompassing tasks like estimating fully or partially missing images~\citep{10087263,9741305}, transforming medical knowledge across modalities~\citep{wang2023attentive,10167641,9758823}, and enhancing image resolution~\citep{wang2023novel,jafari2019echocardiography}.

Recently, deep learning-based methods have been used for medical image synthesis~\citep{yi2019generative}. State-of-the-art deep learning-based methods employ convolutional neural networks (\ie U-net and auto-encoder) to capture implicit dependencies among pixels/voxels during end-to-end training, where the model maps the input image directly to the output image without any intermediate steps~\citep{cao2023autoencoder,zhou2021models,fu2019deep,ronneberger2015u}. 
Generative adversarial networks (GANs) have also been used to improve medical image synthesis results~\citep{wang2023attentive,9758823,yi2019generative,shin2018medical, bermudez2018learning,baur2018generating}. GANs estimate the density function of the data distribution and generate samples consistent with the distribution of the real sample set. Recently diffusion probabilistic models also have been proposed to synthesize medical~\citep{txurio2023diffusion,pinaya2022brain,kazerouni2022diffusion}. The diffusion probabilistic model perturbs input data with Gaussian noise and then learns to reverse the process, recovering the noise-free data from the noisy samples.

However, all the methods mentioned above require specific training on available datasets for image synthesis, which is impractical in data-free scenarios where training data is unavailable during synthesis. To the best of our knowledge, no medical image synthesis method has been proposed that can generate samples without training a separate generative model (\eg, GAN) or adding extra layers (\eg, U-Net).

\subsection{Normalization Layers}
The first introduced normalization layer was Batch Normalization (BN) which was introduced to address the internal covariate shift in the centralized training of deep neural networks~\citep{ioffe2015batch}. Since then, normalization layers have become indispensable to neural network architectures. However, subsequent studies questioned the effect of BN on internal covariate shift and proposed that its main benefits lie in enabling training with larger learning rates and smoothing the optimization surface~\citep{bjorck2018understanding,santurkar2018does}. Additionally, the dependence of BN on the batch size has also been demonstrated as its main drawback. Batch Re-Normalization (BRN)~\citep{ioffe2017batch} has been proposed to solve this issue by normalizing the training data using the running moments estimated by far, achieved through the incorporation of a re-parameterization trick

Spatial normalization layers were introduced to address the discrepancy between training and testing. Among them, Layer Normalization (LN)~\citep{ba2016layer} and Instance Normalization (IN)~\citep{ulyanov2016instance} normalize the input feature along the spatial dimension. However, there are differences in their approaches. LN normalizes the input feature jointly, while IN performs channel-wise normalization. To provide a more versatile solution, Group Normalization (GN)~\citep{wu2018group} was introduced, which serves as a general case of both LN and IN. It starts by dividing the spatial channels into groups and then conducts normalization. It is worth noting that GN becomes equivalent to LN when the number of groups is set to one. Similarly, GN aligns with IN when the number of groups matches the number of channels.
While these spatial normalization layers have successfully addressed certain limitations associated with BN, their effectiveness has been observed to be context-dependent and limited to specific applications.

Subsequent studies proposed the combination of spatial normalization layers and BN, merging their statistics to leverage the benefits of both techniques, such as Switch Normalization (SN)~\citep{luo2018differentiable}, which combine all three IN, LN and BN layers. 
Additionally, various specialized normalization layers have been proposed to tackle scenarios where the test distribution deviates from the training distribution~\citep{bronskill2020tasknorm,wang2019transferable}. 
Research has been conducted to explore the impact of normalization layers in rehearsal-based continual learning methods. 
Continual Normalization (CN)~\citep{pham2022continual} proposed a combination of GN and BN to mitigate the cross-task normalization effect in online rehearsal-based continual learning. Rebalancing Batch Normalization (RBN)~\citep{cha2023rebalancing} addresses the issue of imbalanced numbers of saved data and newly incoming data by rebalancing the computation of normalization over samples from different classes. However, the specific influence of normalization layers in data-free continual learning has yet to be investigated.
\begin{figure*}[h]
	\centering
	\includegraphics[width=0.88\textwidth]{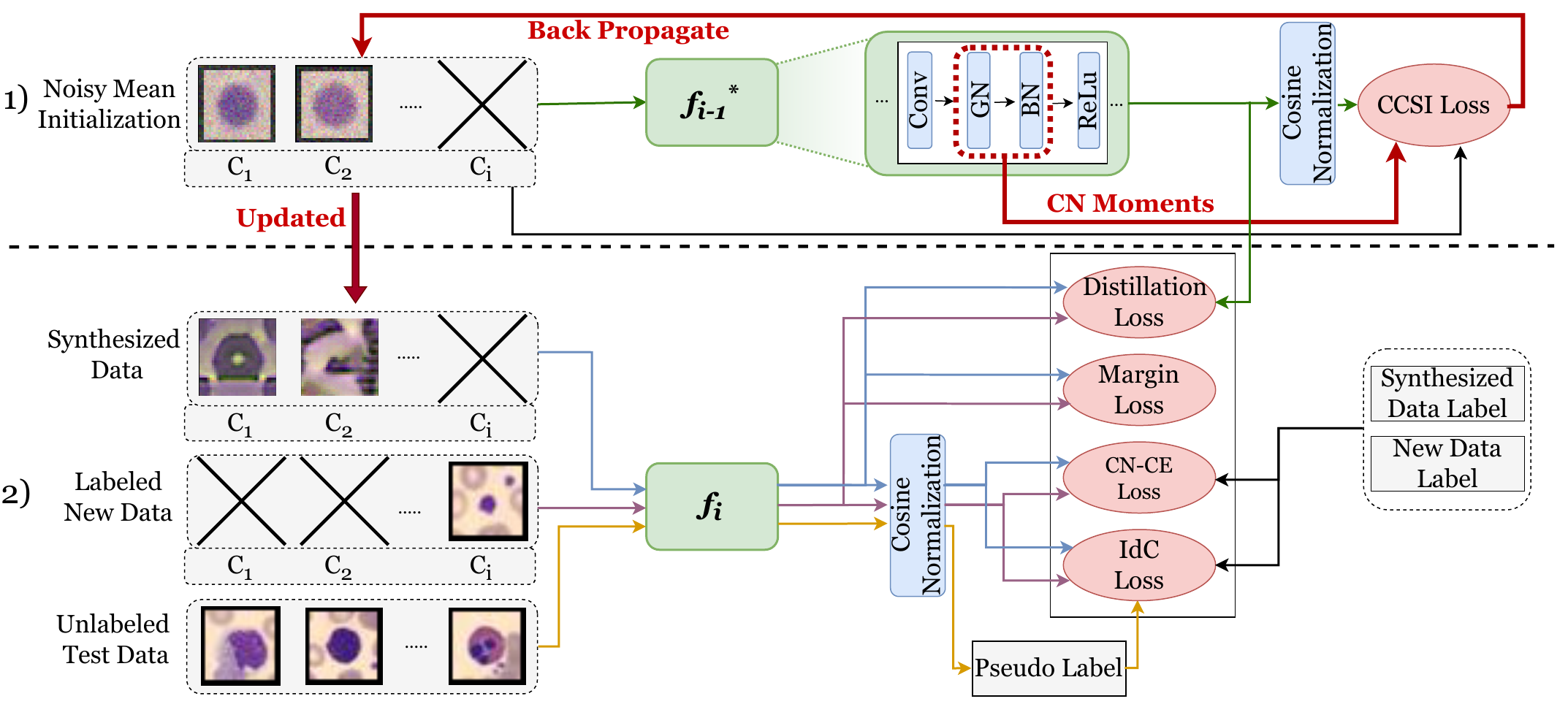}
	\caption{
 The class incremental learning pipeline of \ours. Two main steps of \ours contain: 1) Continual class-specific data synthesis (Sec.~\ref{part1}): Initialize a batch of images with the mean of each class to synthesize images using a frozen model trained on the previous task, $f_{i-1}^{*}$. Update the batch by back-propagating with Eq.~\eqref{eq:CIloss} and using the statistics saved in the CN as a regularization term (Eq.~\eqref{eq:r_cn}); 2) Model update on new tasks (Sec.~\ref{part2}): Leverage information from the previous model using the distillation loss. To prevent catastrophic forgetting of past tasks, we mitigate domain shift between synthesized and original data with a novel intra-domain conservative (IdC) loss (Sec.~\ref{IdC}), a semi-supervised domain adaptation technique and encourage robust decision boundaries and overcome data imbalance with the margin loss (Sec.~\ref{Margin}) and cosine-normalized cross-entropy (CN-CE) loss (Sec.~\ref{CN-CE}). Best viewed in coloured print.
 }
	\label{fig:thresh}
\end{figure*}
\section{Method}
\label{Methods}

This section is structured as follows. We begin by explaining the problem setting of data-free incremental learning in Sec.~\ref{sec:problemsetting}. Next, in Sec.~\ref{sec:overview}, we identify the challenges in the setting and give an overview of our proposed pipeline, emphasizing three crucial factors that drive our design: \emph{1) normalization layer, 2) data synthesis, and 3) loss functions}. Furthermore, we elaborate on the techniques suggested to tackle the first two factors in Sec.~\ref{part1}. Finally, we present the new loss terms utilized to address the third factor in Sec.~\ref{part2}.

\subsection{Problem Setting of Data-free Incremental Learning}
\label{sec:problemsetting}
We start by describing the general problem of class incremental learning, which involves a sequence of classification tasks where new classes are added over time to a prior set of classes. The objective is to maintain high classification accuracy across all tasks. Let us denote $x^t \in \mathcal{X}^t$ as the input image, $y^t\in \mathcal{Y}^t$ as the class label w.r.t. the task $t$, and we have $\{\mathcal{Y}^t\} \subset \{\mathcal{Y}^{t+1}\}$. In class incremental learning, the goal is to control the generalization error of all seen tasks:
\begin{align*}
    \sum_{t=1}^{T} \mathbb{E}_{(x^t,y^t)\sim(\mathcal{X}^t,\mathcal{Y}^t)}\left[ \ell(f_t(x^t;\theta), y^t)\right].
\end{align*}
Given a loss function $\ell$, classifier parameter $\theta$, the current task $T$, and the model $f_{t}$ trained on sub-tasks $t$, when processing a new task $t+1$ under \emph{data-free} setting, there is no access to data $(\mathcal{X}^\tau,\mathcal{Y}^\tau)$ for $\forall \tau \leq t$. 

\subsection{Overview of the Proposed Pipeline and Design Rationals}
\label{sec:overview}

The main challenge in data-free incremental learning is restoring information from old tasks to prevent catastrophic forgetting. Without accessing any data from the old tasks, we propose \ours as a two-step data-free class incremental framework as depicted in Fig.~\ref{fig:thresh}. In the first step, we generate synthetic images representing the old data information distribution from the frozen classification model trained on the previous task. In the second step, we perform class incremental training on the new and synthesized data. There are three key factors in the proposed pipeline:

\paragraph{1) Selection of normalization layer}
Common deep neural network layers may fail to perform well in class-incremental learning settings, especially with non-i.i.d data. Widely used normalization layers (such as BNs) are the most prone to harm performance as their saved statistics in training time are used directly on test data. 
Particularly in our proposed pipeline, as the synthetic data is restored from the trained model, the selection of the normalization layer also affects synthetic data utility.
However, the effect of normalization layers on synthesized data remains largely unexplored in data-free incremental learning settings.  

\paragraph{2) Data-free data synthesis with model inversion}
As synthesized data is used as a representation of previous classes, our pipeline's performance highly relies on the quality of synthesized data.
To obtain the knowledge of
training data from trained model $f_{t-1}^{\star}$ on previous classes, some data-free studies~\citep{ayromlou2022class,choi2021dual,smith2021always} has explored using Deep Inversion~\citep{yin2020dreaming}, which is originally formulated to distill knowledge from a trained model for transfer learning that optimizes following objective function on input $x$ with a randomly initialized $\hat{x}$ and label $y$ pixel-wise: 

\begin{align}
\label{eq:CIloss}
    \min_{\hat{x}}\mathcal{L}_{Class}(\hat{x},y)+\mR(\hat{x}),
\end{align}
where $\mathcal{L}(\cdot)$ is the classification loss used for training and $\mR(\cdot)$ is a regularization term to improve the fidelity of the synthetic images.
However, the existing methods have predominantly focused on the direct deployment of synthesized data using Deep Inversion techniques for preventing catastrophic forgetting in rehearsal-based scenarios over natural images. These approaches have not been extensively adapted to the iterative process of incremental learning. Furthermore, they have not fully explored the potential of synthesizing high-fidelity and utility samples for previously inaccessible classes, which could facilitate effective knowledge transfer in the incremental learning setting over medical images. 

\paragraph{3) Loss functions}
Not only the quality of generated data will affect the final performance, but also the choice of loss functions for training is crucial to decrease the inevitable quality gap between real and synthesized images to have a balanced training among new samples and previous classes. Most of the previous methods focused on rebalancing the classifier in rehearsal-based methods~\citep{hou2019learning}; however, the remaining quality and domain gap between synthesized and real data has been overlooked, which harms the training process.

\paragraph{Remark} \ours{} can improve the design of the above three factors. \emph{First}, we propose to leverage CN to overcome saved biased statistics for data-free incremental learning, which would harm both data synthesis and training step; \emph{Second,} we improve the quality of synthesized data by adopting the CN statistics and class-specific landmarks of medical images in Deep Inversion; \emph{Finally}, we employ three novel losses to a) overcome the distribution shift among synthesized and real data, b) maximize the separation between the distributions of the new and previous classes in the latent space, c) learn features of the new classes while preserving knowledge from prior classes.

\subsection{Continual Class-Specific Impression for Data Synthesis}
\label{part1}

With the unique characteristics of medical images - such as their highly specialized content, structured noise, skewed distribution, and the need for high resolution and precision - the task of synthesizing medical images becomes significantly more challenging.
We have observed that the efficacy of Deep Inversion (Eq.~\eqref{eq:CIloss}) in transferring knowledge via synthesized images in class incremental scenario over medical data is significantly affected by two factors: a) the initialization values of the optimization process ($\hat{x}$), and b) matching the moments of normalization layers as the regularization term ($\mR(\hat{x})$).

\subsubsection{Mean image initialization}
The initialization of generating impressions via optimizing Eq.~\eqref{eq:CIloss} in our proposed pipeline is crucial, as the parameter space is high-dimensional (batch size $\times$ image size). The previous method adopted a random Gaussian for image synthesis of the natural domain~\citep{yin2020dreaming}. However, due to intricate patterns among different classes in medical images, we observed that such a strategy struggles to distinguish synthetic medical images from different classes and even to optimize optimal optimization over $\hat{x}$. To enhance the quality of data synthesis and improve the utility of synthetic data for downstream classification tasks, we propose initializing synthetic image $\hat{x}$ with the averaged image of each class $\bar{x}_k^t$ ($k\in \mathcal{Y}^t$)  for optimizing Eq.~\eqref{eq:CIloss}. This strategy is \emph{uniquely} inspired by the commonality of anatomical landmarks among samples within a medical imaging class, which has been overlooked in previous studies~\citep{yin2020dreaming}.

We note that requiring a mean image from a dataset allows us to impart class-specific impressions without violating privacy regulations such as re-identification and differential privacy~\citep{abadi2016deep}. It is also important to note that the averaged images, as shown in the second row of Fig.~\ref{fig:thresh}, are visually different from the real images shown in the first row of Fig.~\ref{fig:thresh}. This is in line with the aim of our data synthesis, which is to generate images following the distributions of past classes rather than reconstructing training data points, as model inversion attacks do~\citep{hatamizadeh2022gradient,huang2021evaluating,yin2021see}.

\subsubsection{Normalization layers} Matching saved statistics of normalization layers (\eg, the widely BNs in convolutional neural networks~\citep{ayromlou2022class}) with the running statistics during image generation as regularizer in Eq.~\eqref{eq:CIloss}, controls the optimization process. However, using common BN's statistics can be highly biased towards newly introduced classes as they are overwritten by sequentially arrived data with a non-stationary distribution during continual training. 
To overcome these limitations of BN in generating synthetic data, which can further affect the performance of our pipeline, we propose leveraging the recently proposed novel normalization strategy, CN~\citep{pham2022continual} in the data synthesis process. The feature map of each mini-batch for a given normalization layer is represented as $m\in  \mathbb{R}^{B\times C \times D}$ where $B, C, D$ are the batch size,  number of channels, and the dimension of the input vector. CN incorporates both GN and BN as
\begin{align}
\label{eq:CIloss_1}
    m_{CN} \longleftarrow BN(GN(m)),  
\end{align}
in which feature maps are computed across the mini-batch dimension~($B$), feature maps of GN are normalized spatially by incorporating moments of each feature into the normalization process~\citep{wu2018group}. Then, GN divides the channel $C$ into groups, where each group's moments are used in normalization:
\begin{align}
\label{eq:GN_divide}
    m^\prime \in  \mathbb{R}^{B\times G \times K \times D} \longleftarrow m \in  \mathbb{R}^{B\times C \times D} ,\; where\: K=\frac{C}{G}.
\end{align}
GN calculates the moments $\mu_{lg}^G$ and ${\sigma^G_{lg}}^2$ for each group $g$ by averaging over $K$ dimension. 
These moments are independent of the batch size and have better consistency over each class's sample in training and inference. 
CN combines both normalization methods by enjoying the fast convergence induced by BN and alleviating the problem of forgetting moments of the previous class due to overwriting them in incremental training steps using the sample-specific normalization values for each class computed during GN.

\subsubsection{Regularization terms for inversion-based synthesis}
Despite the promising performance of CN in rehearsal-based continual learning scenarios~\citep{pham2022continual}, where old task data is stored in a memory buffer, \emph {how to adapt CN to a data-free setting has remained unexplored}. To address this gap, we build upon our preliminary work on BN-based class impression~\citep{ayromlou2022class} and enhance it using CN. To use the values computed in both GN and BN during training for synthesizing samples of previous classes in our synthesis paradigm, we save the training-time values of GN as ${ \mE(\mu_l^G(x)|\mX),\mE({\sigma_{l}^G}^2(x)|\mX)}$, similar to BN, rather than following \citet{pham2022continual} which calculates the moments of GN separately during training and inference.
Next, we define $\mR_{\rm CN}$ as
 \[ 
   \mR_{\rm CN} =
   \]
   \[
   \sum_l \|\mu^G_l(\hat{x}) - \mE(\mu^G_l(x)|\mX)\|_2 + \sum_l \|{\sigma^G_{l}}^2(\hat{x}) - \mE({\sigma^G_{l}}^2(x)|\mX)\|_2 + 
   \]
   \begin{align}
   \label{eq:r_cn}
   \sum_l \|\mu_l^B(\hat{x}) - \mE(\mu_l^B(x)|\mX)\|_2 + \sum_l \|{\sigma_{l}^B}^2(\hat{x}) - \mE({\sigma_{l}^B}^2(x)|\mX)\|_2 ,
\end{align}
where \{$\mu^G_l$ , ${\sigma_{l}^G}^2$\} and \{$\mu^B_l$ , ${\sigma_{l}^B}^2$\} are the batch-wise mean and variance estimates of feature maps corresponding to the GN and BN of $l$-th CN layer, respectively. 
It is noteworthy that, compared to using BN statistics, the proposed CN (Eq.~\eqref{eq:r_cn}) not only overcomes the batch stability issue by incorporating each feature’s statistics as \{$\mu^G_l$ , ${\sigma_{l}^G}^2$\} into its normalization to have sample-specific normalization but also leverage them as additional feature statistics to supervise the data synthesis. 

Following~\citep{yin2020dreaming}, we also add $\mR_{\rm TV}(\cdot)$ and  $\mR_{\ell_2}(\cdot)$ as regularizers to penalize the total variance and the $\ell_2$ norm of the generated image batch $\hat{x}$, respectively and define regularization term ($\mR(\hat{x})$) as
\begin{align}
\label{eq:reg}
    \mR(\hat{x})=\alpha_{\rm tv}\mR_{\rm TV}(\hat{x}) + \alpha_{\ell_2}\mR_{\ell_2}(\hat{x}) + \alpha_{\rm cn}\mR_{\rm CN}(\hat{x},\mathcal{X}).
\end{align}
where $\alpha$s are scaling factors to balance the importance of each regularizer. Finally, starting from the mean image of each class, we optimize Eq.~\eqref{eq:CIloss} as our final data synthesis objective.

\subsection{Classification Losses}
\label{part2}
\begin{figure}[t]
	\centering
 \subfigure[ \textcolor{black}{Intra-domain Contrastive Loss}]
        {
        \includegraphics[width=0.8\columnwidth]{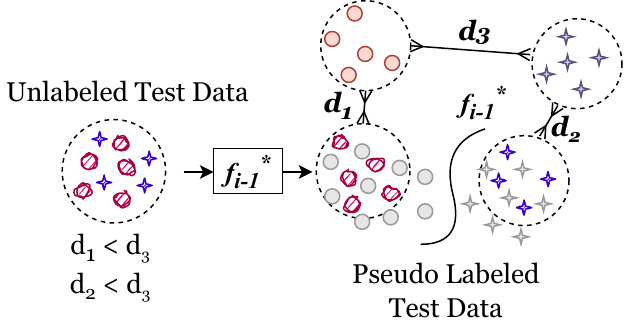}
        \label{fig:first_sub}
        }
	
      \subfigure[Cosine-Normalized Cross-Entropy Loss]
    {
        \includegraphics[width=0.4\columnwidth]{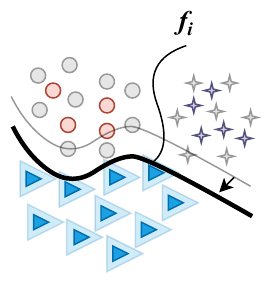}
        \label{fig:first_sub}
    }
        \hspace{35pt}
         \subfigure[Margin Loss]
        {
        \includegraphics[width=0.4\columnwidth]{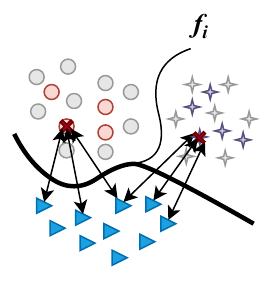}
        \label{fig:first_sub}
        }

	\caption{The effect of each loss on the model's latent space. $f^*_{i-1}$ is the model trained on previous data, and $f_{i}$ is the updated model with new classes. (a) The intra-domain contrastive loss reduces the domain shift by minimizing the distance between the synthesized and test data of the same class and maximizing the distance between synthesized data from different classes. (b) The margin loss enforces the separation between the latent representation of the new class and the previously trained classes. (c) The cosine-normalized cross-entropy loss balances the importance of the new class against the previously trained classes in the latent space to achieve clear class boundaries. Best viewed in coloured print. }
	\label{losses}
\end{figure}

 To enhance the effectiveness of our synthetic data approach in \ours{}, we propose to employ three novel losses, including two regularization terms, the \emph{intra-domain contrastive loss} to overcome the distribution shift among synthesized and real data and the \emph{margin loss} to maximize the separation between the distributions of the new and previous classes in the latent space and a modified bias-correction classification loss, the \emph{cosine-normalized cross-entropy loss}, to balance incremental training. These losses help improve the utility of our synthetic data approach by updating class weights and reducing mismatches between the old representation space and the updated one. The impact of each loss on the training process is shown in Fig.~\ref{losses}.

\subsubsection{Intra-domain contrastive loss}
\label{IdC}
When directly applying synthetic medical images for rehearsal, the resulting classification performance may not generalize well to the original data due to the inevitable distribution mismatch between real and synthetic data. To handle this challenge, we adopt a semi-supervised domain adaptation approach~\citep{singh2021clda} to align the source and target domains. To achieve this domain alignment, we utilize the intra-domain contrastive (IdC) loss, which encourages samples of the same class in both the source and target domains to be closer together while pushing samples of different classes apart. This process is illustrated in Fig.~\ref{losses} (a). In this context,  the source domain ($\mS$) refers to the synthetic images, while the target domain ($\mT$) represents the unlabeled testing data in the new task.

We calculate the centroid $c_k^\mS$ for each class $k$ in the source domain as the mean of the feature representations of all the images in that class. Similarly, we calculate the centroid $c_k^\mT$ in the target domain using the pseudo labels assigned to the samples.
Finally, IdC loss is defined as follows (at task $t$):

    \[ 
    \mL_{\rm IdC} (c_k^\mS,c_k^\mT) =
    \]
    \begin{align}
    \label{eq:contras}
    -\log \frac{\exp(\tau\langle \bar{c}_k^\mS, \bar{c}_k^\mT \rangle)}{\exp(\tau\langle \bar{c}_k^\mS, \bar{c}_k^\mT \rangle) + \sum_{\substack{j=1 \\ \mathcal{Q}\in\{\mS,\mT\}}}^{|\mathcal{Y}^{t-1}|}\mathbbm{1}_{\{j\neq k\}}\exp(\tau\langle \bar{c}_k^\mathcal{Q}, \bar{c}_k^\mT \rangle) },
\end{align}
where $\langle\cdot,\cdot\rangle$ measures the cosine similarity between two vectors and $\tau$ is the temperature hyperparameter in $\exp(\cdot)$ .

\subsubsection{Margin loss}
\label{Margin}
Since each synthetic data point in our pipeline represents a latent space distribution of an inaccessible previous class, repelling new samples from these representations is crucial, as shown in Fig.~\ref{losses} (b). Representation overlapping is a prevalent issue in class incremental learning~\citep{zhu2021class}. This approach contributes to establishing more distinct classification boundaries among different classes.
Previously utilized in the few-shot learning~\citep{li2020boosting} and non-data-free class incremental learning~\citep{hou2019learning}, the margin loss involves comparing distances between sample representations using a specified margin. 
In data-free incremental learning, it becomes even more challenging to learn a good decision that can generalize well across both old and new classes due to the limited model capacity when classes increase and the lack of comprehensive data. To improve the decision boundary, we are the first to apply margin loss in data-free class incremental learning to promote the separation of the new class decision boundary from the old classes. At task $t$, the margin loss is defined as follows:
\begin{align}
\label{eq:margin}
\mL_{\rm margin} = \sum_{k=1}^{|\mathcal{Y}^{t-1}|}\max \left( m - \langle \bar{\theta} , \bar{f}(x) \rangle + \langle \bar{\theta}^k,\bar{f}(x)\rangle, 0 \right),
\end{align}
where $x$ represents the synthesized images as anchors of the previous task's distribution classes, $m$ is the margin tolerance, $\bar{\theta}$ is the embedding vector of $x$'s true class, and $\bar{\theta}^k$ represents the embedding vector of the new class viewed as negatives for $x$. A larger margin promotes a greater separation.

\subsubsection{Cosine-normalized cross-entropy loss}
\label{CN-CE}
Although introduced intra-domain contrastive loss and margin loss can increase the utility of synthesized samples and regularize the feature space, achieved imbalanced class distribution among real and synthesized data can still result in larger embedding magnitudes, leading to a bias toward new classes compared to older ones, as illustrated in Fig.~\ref{losses} (c). In class incremental learning, classifier bias is a common issue arising due to unequal availability or data quality in new classes compared to older ones~\citep{zhu2021class}. 
Cosine normalization has been used in various vision tasks to mitigate this bias caused by magnitude differences~\citep{gidaris2018dynamic,hou2019learning,qi2018low}. To mitigate classifier bias in class incremental learning, we adopt the \textit{Cosine-Normalized Cross-entropy} (CN-CE) loss introduced in \citet{hou2019learning} instead of the common softmax operation used in the classical classification setting. 
While the softmax operation predicts the probability of class $k$ as $p_i(x) = \exp(\theta_i^{\top}f(x)+b_k)/\sum_j\exp(\theta_j^{\top}f(x)+b_i)$, where $f(\cdot)$ is the feature extractor and $\theta_k$, and $b_k$ are the class embedding weights that later are used to calculate cross entropy (CE) loss, CN-CE loss penalizes incorrect predictions and normalizes for magnitude differences between past and new classes embeddings at the current task $t$ as 
\begin{align}
\label{eq:cosine}
\mL_{\rm CN-CE} = -\sum_{k=1}^{|\mathcal{Y}^t|}y_k\log\left(\frac{\exp(\eta\langle \bar{\theta}_k,\bar{f}(x)\rangle )}{\sum_j\exp(\eta\langle \bar{\theta}_j,\bar{f}(x)\rangle )}\right),
\end{align}
where $\bar{v}=v/|v|_2$ is the unit-normalized vector, $\langle\cdot,\cdot\rangle$ measures the cosine similarity between two vectors, and $\eta$ is a temperature hyperparameter.

In sum, the combined loss function used to train the model to recognize new classes while preserving the old knowledge is:
\[
\mL_{\ total} = \mL_{\rm CN-CE} + 
\]
\begin{align}
\label{eq:total}
\alpha_{\rm dist} \mL_{\rm dist} + \alpha_{\rm IdC}\mL_{\rm IdC}+ \alpha_{\rm margin}\mL_{\rm margin},
\end{align}
where $\alpha$s are adjustable scaling factors, and the distillation loss $\mL_{\rm dist}$ is commonly used in class incremental learning~\citep{hou2019learning,li2017learning}. The distillation loss is expressed as $1 - \langle \bar{f}^*(x), \bar{f}(x) \rangle$, where $\bar{f}^*(x)$ is the output of the old model trained on previous tasks and $\bar{f}(x)$ is the output of the new model. The distillation loss helps the new model leverage information from the old model and maintain its performance on previous tasks.
\section{Experiments}
\subsection{Datasets}

\textcolor{black}{Our work specifically focuses on medical image disease classification. Unlike natural datasets, medical datasets typically involve fewer classes but higher pixel counts. To assess the effectiveness of our method, we aimed to select datasets with numerous classes and higher resolutions. Therefore, we chose the publicly available MedMNIST benchmark datasets for continual disease classification, which include various medical image modalities, as well as a more complex, real-world in-house Heart Echo dataset.}

\begin{table}
\caption{Class division for MedMNIST and Heart Echo datasets. In each task, we introduce different classes than previously learned tasks. Our goal is to have a model performing well in all of the introduced classes. }
\begin{center}
\begin{tabular}{l c c c }
\hline
Dataset& Classes & Tasks& Classes Per Task\\
\hline\hline
BloodMNIST & 8 & 4 &  [2,2,2,2] \\
\hline
PathMNIST & 9 & 4 & [3,2,2,2] \\
\hline
OrganaMNIST & 11 & 4 & [3,3,3,2]\\
\hline
TissueMNIST& 8 & 4 & [2,2,2,2]\\
\hline
Heart Echo Dataset & 5 & 4 & [2,1,1,1]\\
\hline
\end{tabular}
\end{center}
\label{tab:distribution}
\end{table}

\subsubsection{MedMNIST}
\begin{figure*}[ht]
	\centering
		\subfigure[BloodMNIST]
		{
			\includegraphics[width=0.38\columnwidth]{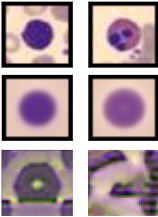}
			\label{fig:first_sub}
		}
		\hfill
		\subfigure[PathMNIST]
		{
			\includegraphics[width=0.38\columnwidth]{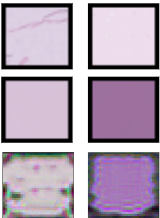}
			\label{fig:first_sub}
		}
		\hfill
		\subfigure[OrganaMNIST]
		{
			\includegraphics[width=0.38\columnwidth]{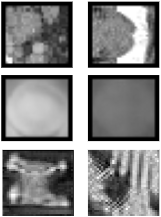}
			\label{fig:first_sub}
		}
        \hfill
		\subfigure[TissueMNIST]
		{
			\includegraphics[width=0.38\columnwidth]{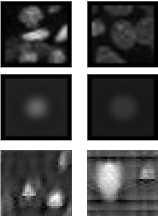}
			\label{fig:first_sub}
		}
        \hfill
		\subfigure[\textcolor{black}{Heart Echo}]
		{
			\includegraphics[width=0.38\columnwidth]{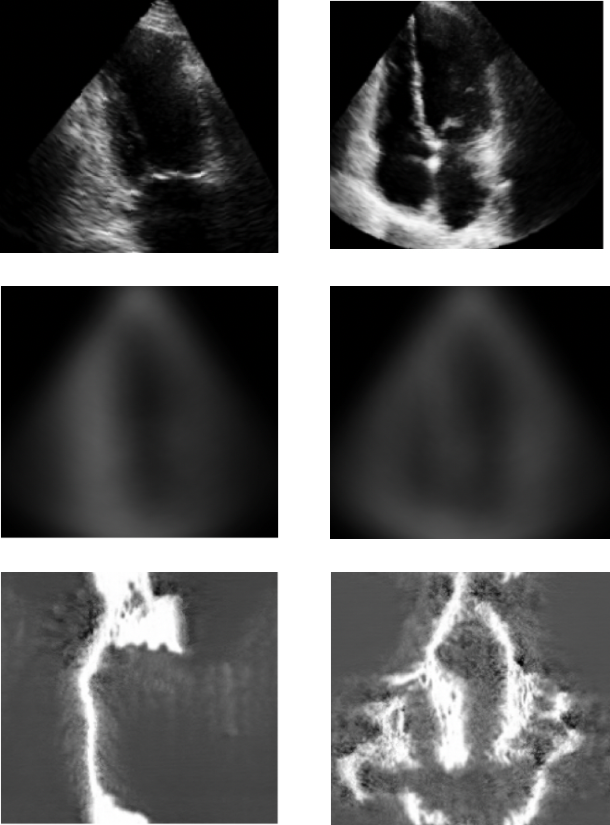}
			\label{fig:first_sub}
		}
		
		\caption{Datasets' samples. Each dataset's first row shows samples from two different classes, the second row is the mean initialization of the respective class, and the third row is the synthesized images. Best view in coloured.}
		\label{fig:sample_table}
	\end{figure*}

The MedMNIST dataset~\citep{medmnist} is a benchmark dataset for medical imaging classification, which has various 2D and 3D modalities. Following the \citet{derakhshani2022lifelonger}, as the benchmark for continual disease classification, we select four modalities with the higher number of diseases that align more with the class incremental setting, as our primary datasets to show the effectiveness of \ours, named BloodMNIST~\citep{acevedo2020dataset}, OrganaMNIST~\citep{bilic2023liver}, PathMNIST~\citep{kather2019predicting}, and TissueMNIST~\citep{bilic2023liver}. The original size of images for all four datasets is $28\times28$, but we pad them to have $32\times32$ images. We split BloodMNIST, OrganaMNIST, PathMNIST and TissueMNIST to separate train, validation and test sets with the ratio of (70-10-20)\%, (60-10-30)\%, (84-10-6)\%, and (70-10-20)\% respectively. Some samples of each dataset are illustrated in the first row of Fig.~\ref{fig:sample_table}.

\begin{table*}
\caption{Conceptual comparison of \ours and state-of-the-art class incremental learning methods. \ours is a data-free approach that synthesizes data without pre-training using additional data. Moreover, we leverage the synthesized data to retain the memory of previous classes, address inter-task confusion, and overcome task-recency bias.}
\label{tab:comp_algo}
\begin{center}
\begin{tabular}{l c c c c c}
\hline
\multirow{2}{*}{Method} & \multirow{2}{*}{No pre-training}  & \multirow{2}{*}{Data-free} & Memory  & Inter-task & Bias \\
& &  & preservation & separation &  correction\\
\hline\hline
iCarL~\citep{rebuffi2017icarl} & \cmark & \xmark & \cmark & \xmark & \xmark \\
\hline
\textcolor{black}{GEM~\citep{lopez2017gradient}}& \textcolor{black}{\cmark} & \textcolor{black}{\xmark} & \textcolor{black}{\cmark} & \textcolor{black}{\xmark} & \textcolor{black}{\xmark} \\
\hline
\textcolor{black}{AGEM~\citep{chaudhry2018efficient}} & \textcolor{black}{\cmark} & \textcolor{black}{\xmark} & \textcolor{black}{\cmark} & \textcolor{black}{\xmark} & \textcolor{black}{\xmark} \\
\hline
\textcolor{black}{DER~\citep{buzzega2020dark}} & \textcolor{black}{\cmark} & \textcolor{black}{\xmark} & \textcolor{black}{\cmark} & \textcolor{black}{\xmark} & \textcolor{black}{\xmark} \\
\hline
\textcolor{black}{DER++~\citep{buzzega2020dark}} & \textcolor{black}{\cmark} & \textcolor{black}{\xmark} & \textcolor{black}{\cmark} & \textcolor{black}{\xmark} & \textcolor{black}{\xmark} \\
\hline
LUCIR~\citep{hou2019learning} & \cmark & \xmark & \cmark &\cmark & \cmark\\
\hline
\textcolor{black}{SI~\citep{zenke2017continual}} & \textcolor{black}{\cmark} & \textcolor{black}{\cmark} & \textcolor{black}{\xmark}  & \textcolor{black}{\xmark} & \textcolor{black}{\xmark} \\
\hline
\textcolor{black}{oEWC~\citep{schwarz2018progress}} & \textcolor{black}{\cmark} & \textcolor{black}{\cmark} & \textcolor{black}{\xmark}  & \textcolor{black}{\xmark} & \textcolor{black}{\xmark} \\
\hline
LwF~\citep{li2017learning}  & \cmark & \cmark  & \xmark &  \xmark & \xmark\\
\hline
CLBM~\citep{yang2021continual} & \xmark & \cmark & \cmark  & \xmark & \xmark \\
\hline
\textcolor{black}{GPM~\citep{saha2021gradient}} & \textcolor{black}{\cmark} & \textcolor{black}{\cmark} & \textcolor{black}{\cmark}  & \textcolor{black}{\xmark} & \textcolor{black}{\cmark} \\
\hline
ABD~\citep{smith2021always} & \cmark & \cmark & \cmark & \cmark & \xmark \\
\hline
\textbf{\ours (ours)} & \cmark & \cmark & \cmark & \cmark & \cmark \\
\hline
\end{tabular}
\end{center}
\label{tab:components}
\end{table*}

\subsubsection{Heart Echo Data}
We use the Heart Echo dataset as our second dataset to evaluate \ours over a more realistic and complex dataset with a higher resolution.  As we do pixel-wise optimization, increasing the resolution of generated images may be challenging. Heart Echo is a private dataset of 11,062 cines (videos) of 2151 unique patients diagnosed with various heart diseases with a size of $256\times256$, randomly chosen from a hospital picture archiving system with authorization. The cines were captured by six devices containing an average of 48 frames per cine. An experienced cardiologist labelled the cines into five different views, with the distribution shown in Table~\ref{tab:distribution}. We perform a frame-level classification by splitting frames into training, validation, and test sets based on the subject with a ratio of (70-10-20)\%.

\subsection{Experimental Settings}
\label{sec:exp}

\begin{figure*}[ht]
	\centering
    \subfigure[BloodMNIST]
        {
        \includegraphics[width=0.6\columnwidth]{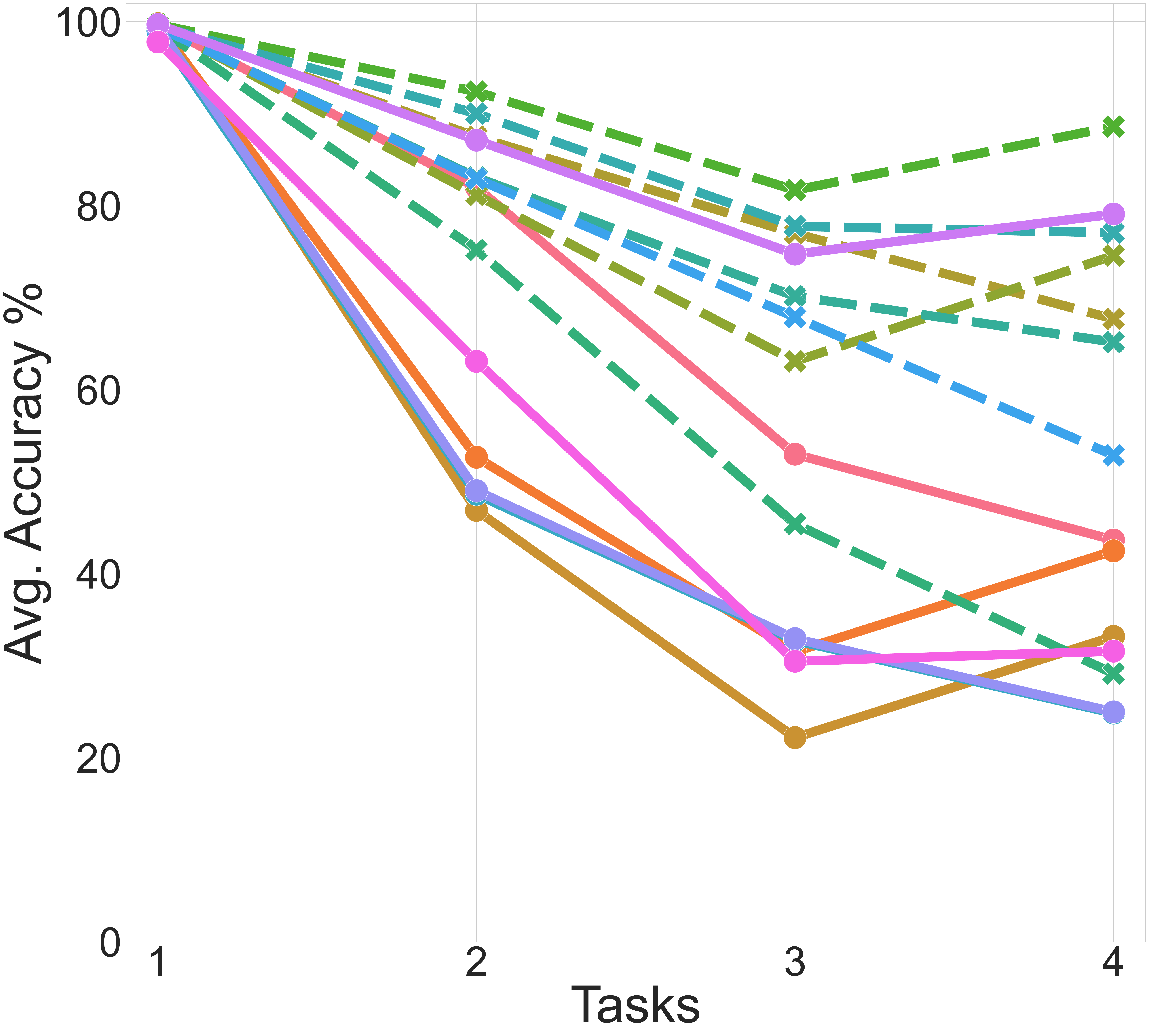}
        \label{fig:SOTAblood}
        }
        \hspace{2pt}
        \subfigure[PathMNIST]
        {
        \includegraphics[width=0.6\columnwidth]{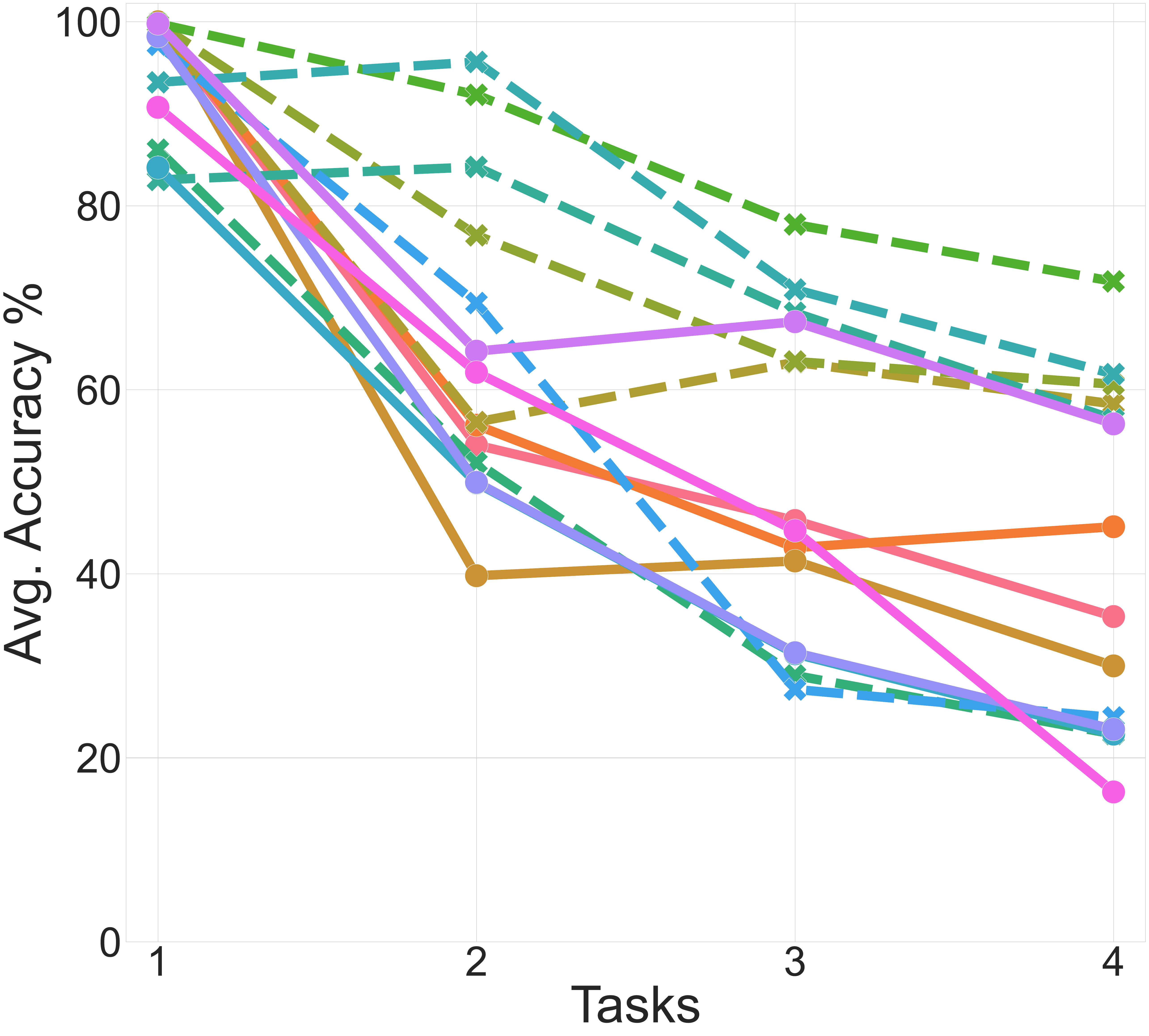}
        \label{fig:SOTApath}
        }
        \hspace{2pt}
        \subfigure[OrganaMNIST]
        {
        \includegraphics[width=0.6\columnwidth]{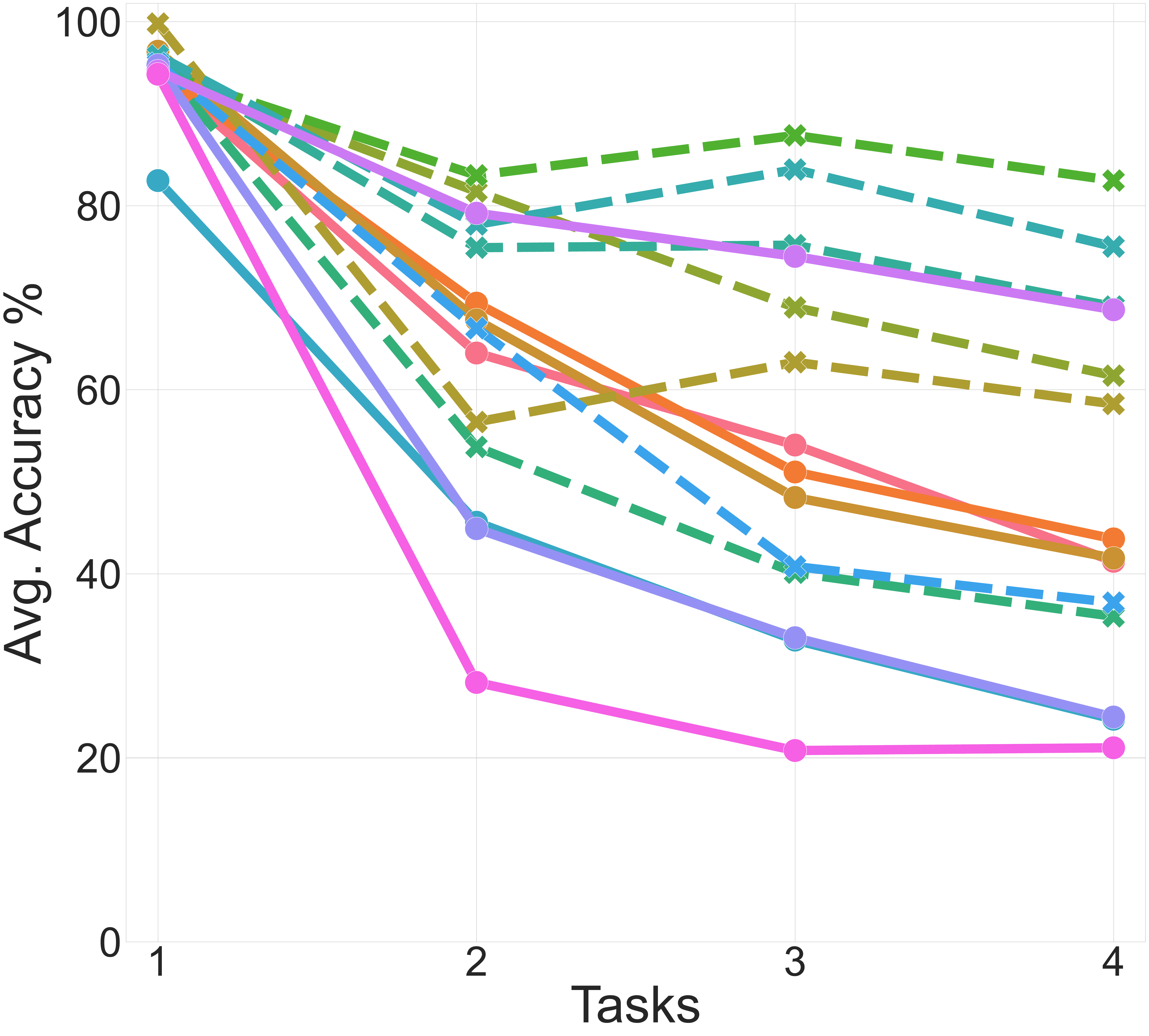}
        \label{fig:SOTAOrgana}
        }
        
        \subfigure[TissueMNIST]
        {
        \includegraphics[width=0.6\columnwidth]{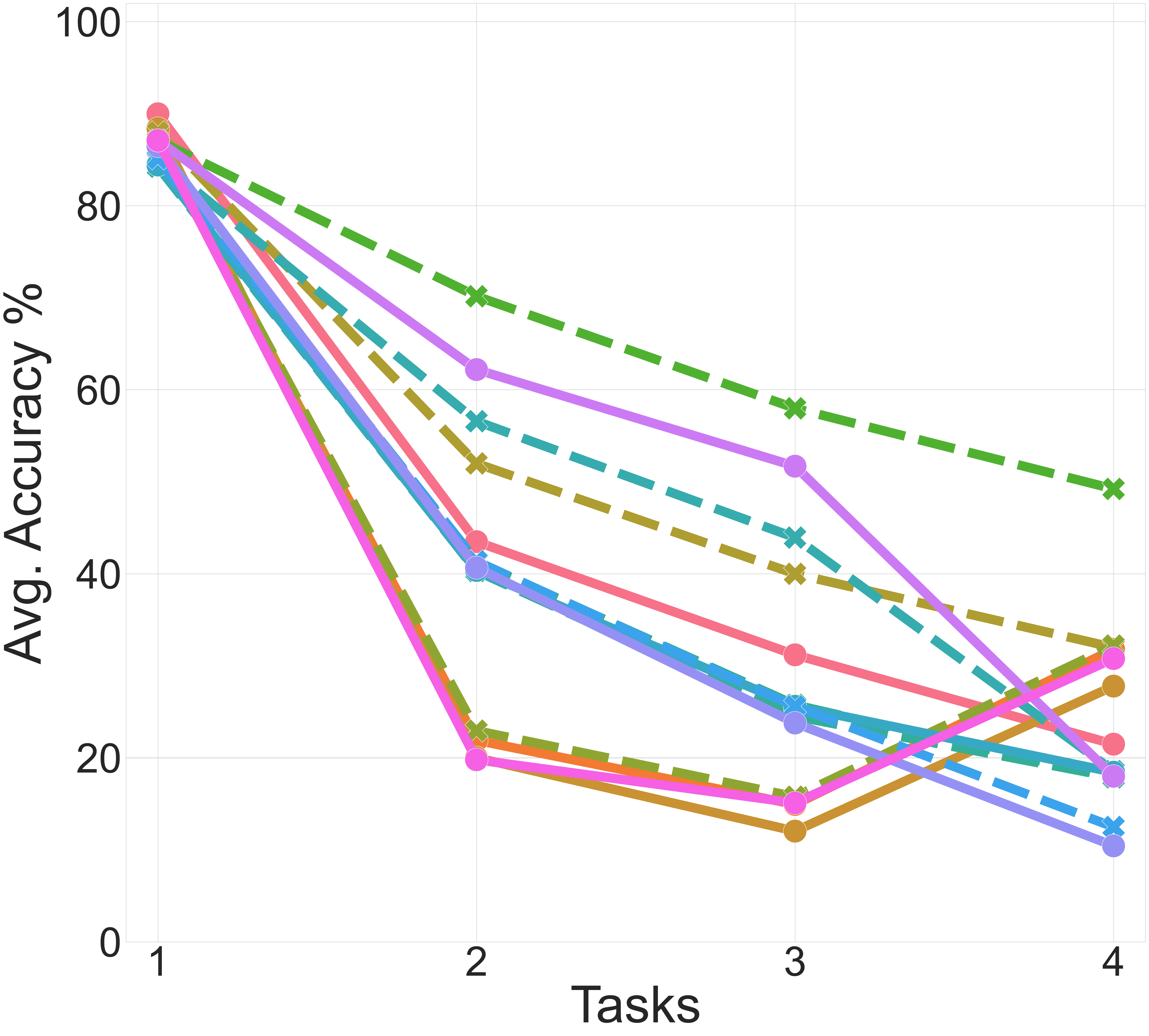}
        \label{fig:SOTATissue}
        }
        \hspace{2pt}
         \subfigure[Heart Echo]
        {
        \includegraphics[width=0.6\columnwidth]{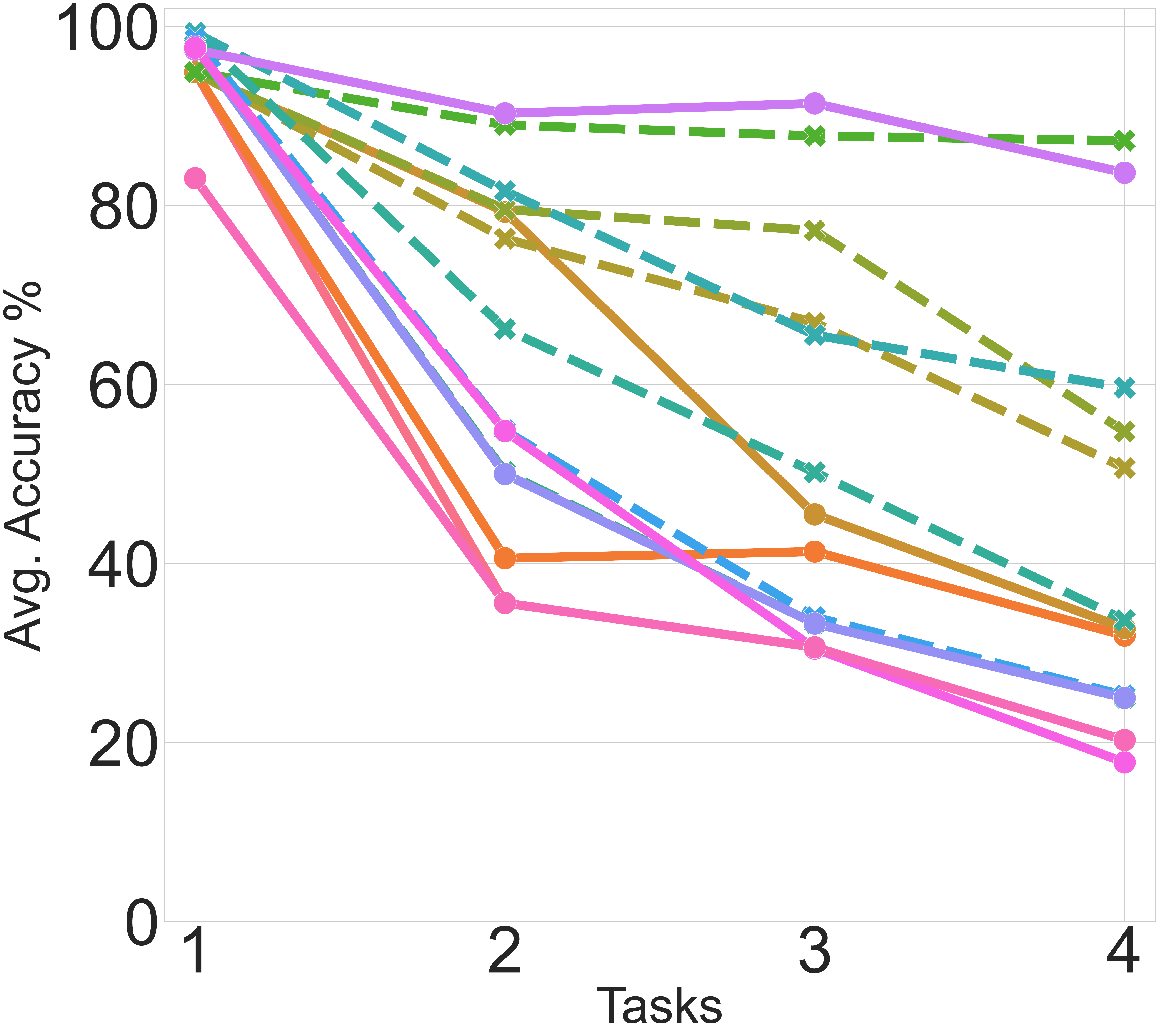}
        \label{fig:SOTAHeart}
        }
        \hspace{23pt}
         \subfigure
        {
        \includegraphics[width=0.5\columnwidth]{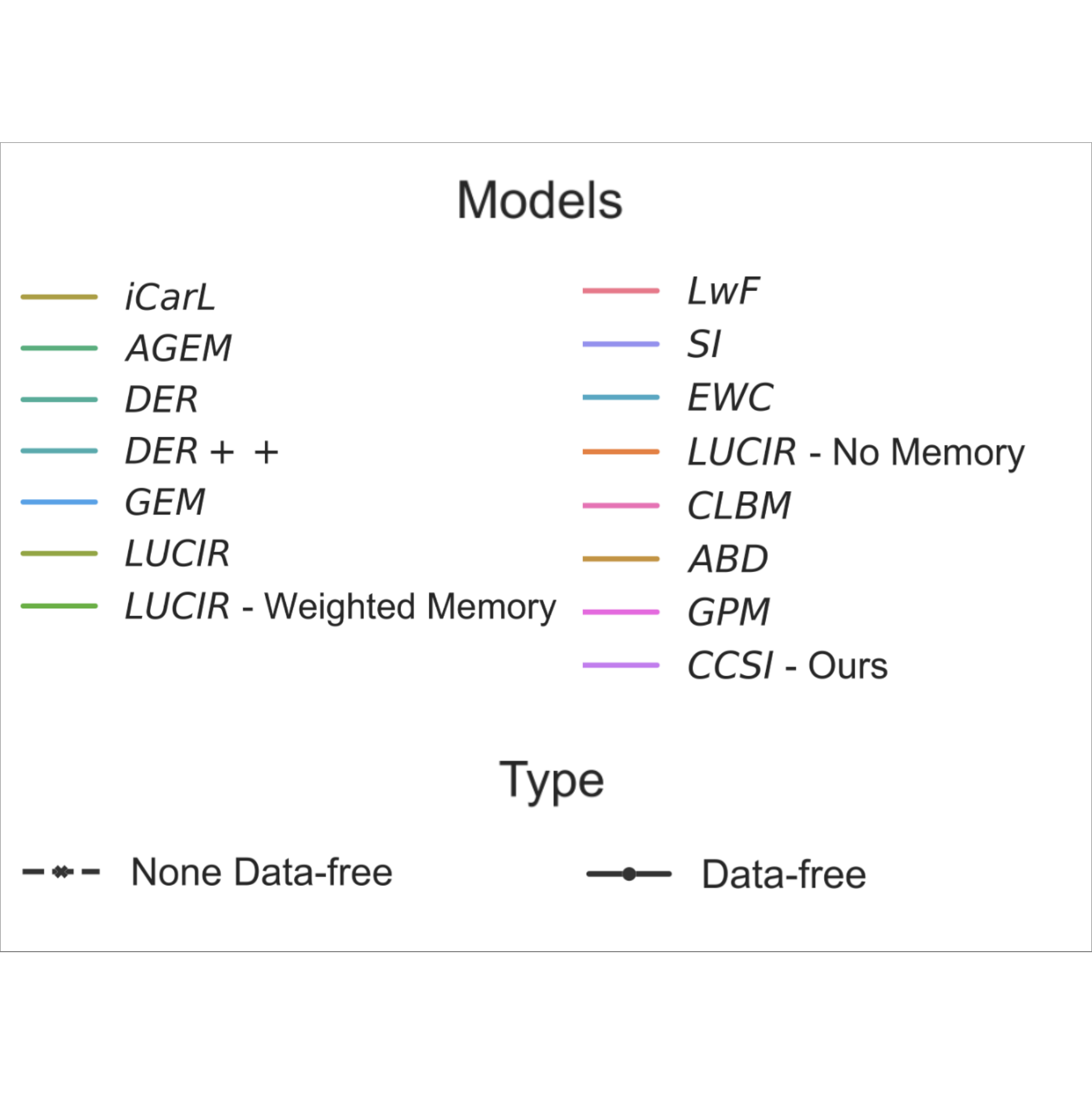}
        \label{fig:sotalegend}
        }
	\caption{Testing accuracies on all tasks compared with state-of-the-art class-incremental learning. \textcolor{black}{Dashed lines represent non-data-free methods, while straight lines represent data-free methods. We outperform all data-free methods on all datasets except TissueMNIST. While we surpass some non-data-free methods, we achieve comparable results with others.} }
	\label{fig:SOTA}
\end{figure*}

Table~\ref{tab:distribution} shows the class division for incrementally introducing in all datasets. Since medical datasets have fewer data points compared to natural image benchmark datasets, we use a ResNet-based~\citep{he2016deep} CNN with three residual blocks of two layers each and replace BN with CN for the classification model in \ours. The models are implemented with PyTorch and are trained on one NVIDIA Tesla V100 GPU with 16GB of memory. As outlined in Sec.~\ref{Methods}, our implementation consists of two main steps for each task. In the image synthesis step, we freeze the model weights and use the Adam optimizer to optimize a batch of 40 images for 2000 epochs with a learning rate of 0.01. In the incremental learning step with a new class, we update the model weights using the SGD optimizer and continue learning for 100 epochs with a batch size of 40. We use Weights \& Biases Sweeps~\citep{wandb}, an automated hyperparameter optimization tool, to search for optimal hyperparameters for synthesizing images.

\subsection{Comparison with Baselines}
\label{sec:baseline}

\begin{table*}[t]
\centering

\caption{Testing accuracies of final task over MedMNIST and Heart Echo datasets compared with \textcolor{black}{data-free} baselines of class-incremental learning. 
\ours shows consistently higher accuracy, up to 51\% increase compared to the state-of-the-art data-free methods.
}
\begin{center}
\resizebox{\linewidth}{!}{%
\begin{tabular}{l c c c c c c}
\hline
\multirow{2}{*}{Method} & \multirow{2}{*}{Coreset Data} & \multicolumn{5}{c}{Accuracy over all tasks}\\
\cline{3-7}
 & & BloodMNIST & PathMNIST & OrganaMNIST & TissueMNIST & Heart Echo\\
\hline\hline

\textcolor{black}{SI~\citep{zenke2017continual}} & \textcolor{black}{No} & \textcolor{black}{25.00} & \textcolor{black}{23.11} & \textcolor{black}{24.44} & \textcolor{black}{15.73} & \textcolor{black}{25.00}\\

\textcolor{black}{oEWC~\citep{schwarz2018progress}} & \textcolor{black}{No} & \textcolor{black}{24.96} & \textcolor{black}{24.20} & \textcolor{black}{24.20} &\textcolor{black}{18.44}& \textcolor{black}{25.04}\\

LwF~\citep{li2017learning} & No & \underline{43.68} & 35.36 & 41.36 & 21.49 & 20.29\\

LUCIR~\citep{hou2019learning} & No & 42.50 & \underline{45.12} & \underline{43.80} & \textbf{31.80} & 31.96\\

CLBM~\citep{yang2021continual}  & Prior Memory & NA & NA & NA & NA & 20.29\\

\textcolor{black}{GPM~\citep{saha2021gradient}} & \textcolor{black}{Projected Memory} & \textcolor{black}{31.60} & \textcolor{black}{16.30} & \textcolor{black}{21.10} & \textcolor{black}{\underline{30.80} }&\textcolor{black}{ 17.80}\\

ABD~\citep{smith2021always} & Synthesized & 33.20 & 30.00 & 41.72 & 27.30 & \underline{32.78}\\
\hline
\ours (ours) & Synthesized & \textbf{79.10} & \textbf{56.30} &\textbf{68.70}& 28.00 & \textbf{83.66}\\

\hline
\end{tabular}
}
\end{center}
\label{tab:results_data_free}
\end{table*}

\begin{table*}[t]
\centering

\caption{Testing accuracies of final task over MedMNIST and Heart Echo datasets compared with \textcolor{black}{non-data-free} baselines of class-incremental learning. 
}
\begin{center}
\resizebox{\linewidth}{!}{%
\begin{tabular}{l c c c c c c}
\hline
\multirow{2}{*}{Method} & \multirow{2}{*}{Coreset Data} & \multicolumn{5}{c}{Accuracy over all tasks}\\
\cline{3-7}
 & & BloodMNIST & PathMNIST & OrganaMNIST & TissueMNIST & Heart Echo\\
\hline\hline

\textcolor{black}{AGEM~\citep{chaudhry2018efficient} } & \textcolor{black}{Saved Memory} & \textcolor{black}{29.15} & \textcolor{black}{22.59} & \textcolor{black}{35.58} & \textcolor{black}{17.83} & \textcolor{black}{25.00}\\

\textcolor{black}{GEM~\citep{lopez2017gradient}} & \textcolor{black}{Saved Memory} & \textcolor{black}{ 52.88} & \textcolor{black}{24.39} & \textcolor{black}{36.87} & \textcolor{black}{12.49} & \textcolor{black}{25.27}\\

iCarL~\citep{rebuffi2017icarl} & Saved Memory & 67.70 & 58.46 & 63.02 & 32.00 & 50.68\\


\textcolor{black}{DER~\citep{buzzega2020dark}} & \textcolor{black}{Saved Memory} & \textcolor{black}{65.19} & \textcolor{black}{56.8} & \textcolor{black}{69.05} & \textcolor{black}{17.90} & \textcolor{black}{33.69}\\

\textcolor{black}{DER++~\citep{buzzega2020dark}} & \textcolor{black}{Saved Memory} & \textcolor{black}{77.07} & \textcolor{black}{\underline{61.71}} & \textcolor{black}{\underline{75.56}} & \textcolor{black}{18.56} & \textcolor{black}{59.63}\\
LUCIR~\citep{hou2019learning} & Saved Memory & 74.59 & 61.56 & 61.54 & \underline{32.21} & 54.75\\

LUCIR~\citep{hou2019learning} & Weighted Memory & \textbf{88.60} & \textbf{71.78} & \textbf{82.74} & \textbf{49.23} & \textbf{87.24}\\

\hline
\ours (ours) & Synthesized & \underline{79.10} & 56.30 & 68.70 & 28.00 & \underline{83.66}\\

\hline
\end{tabular}
}
\end{center}
\label{tab:results_non_data_free}
\end{table*}

We evaluate \ours a data-free approach against the top-performing baselines in incremental learning on our datasets. These baselines include 1) data-free methods as LWF~\citep{li2017learning}, \textcolor{black}{SI~\citep{zenke2017continual}}, \textcolor{black}{oEWC~\citep{schwarz2018progress}}, CLBM~\citep{yang2021continual}, \textcolor{black}{GPM~\citep{saha2021gradient}}, and ABD~\citep{smith2021always}, 2) non-data-free methods as iCarL~\citep{rebuffi2017icarl}, \textcolor{black}{GEM~\citep{lopez2017gradient}}, \textcolor{black}{AGEM~\citep{chaudhry2018efficient}}, \textcolor{black}{DER ~\citep{buzzega2020dark}}, \textcolor{black}{DER++ ~\citep{buzzega2020dark}}, and LUCIR~\citep{hou2019learning}. Table~\ref{tab:comp_algo} compares feature of these state-of-the-art methods and \ours{}. The results are shown in Table~\ref{tab:results_data_free}, Table~\ref{tab:results_non_data_free}  and Fig.~\ref{fig:SOTA}. Also, some samples of synthesized data with \ours for MedMNIST are illustrated in the third row of Fig.~\ref{fig:sample_table}.

\noindent\textcolor{black}{\textbf{SI}~\citep{zenke2017continual} robustly pulls back more influential network parameters towards a reference weight, which demonstrated proficient performance in past tasks. However, this limitation regarding weights could hinder the network from learning new classes effectively. Consequently, \ours outperforms it with a significant margin across all datasets.}

\noindent\textcolor{black}{\textbf{oEWC}~\citep{schwarz2018progress} conducts distillation into the knowledge module to protect any previously acquired skills, after the acquisition of a new task using the active module. While this progress and comparison strategy is data-free, it fails to effectively transfer knowledge from previous tasks in a class incremental learning scenario. Although it has comparable performance with SI, \ours outperforms it on all datasets, maintaining an approximately 13\% to 85\% advantage. }

\noindent\textbf{LwF}~\citep{li2017learning} uses distillation-based training to mitigate catastrophic forgetting across tasks. It is designed for continual learning scenarios where classifications are performed separately for each task and do not consider the problem of imbalance among classes in different tasks to maximize distinction among intra-task classes. Due to these limitations, \ours outperforms LWF across all datasets, with a gap of a minimum of 6.6\% and a maximum of 63.3\% in the final task over TissueMNIST and Heart Echo datasets, respectively.

\begin{table*}[t]
\caption{Testing accuracy of ablation studies done on different configurations of the proposed framework. These experiments are divided into two main categories: 1) Modifications over the synthesis Step; 2) Modifications over the training Step.}
\begin{center}
\begin{tabular}{l l c c c c}
\hline
\multicolumn{6}{c}{BloodMNIST Dataset}\\
\hline
\multirow{2}{*}{Ablation Type}&
\multirow{2}{*}{Ablation Part} & \multicolumn{4}{c}{Number of Classes (Accuracy)}\\
\cline{3-6}& & 2 & 4 & 6 & 8  \\
\hline\hline
Synthesis Step & w/o synthesis  & 99.70 & 80.60 & 65.16 & 68.60\\
Synthesis Step & w/o regularization  & 99.70 & 79.90 & 34.30 & 37.91\\
Synthesis Step &  BN  & 99.80 & 82.53 & 68.05 & 72.90\\
Synthesis Step & w/o mean initialization & 99.70 & 80.20 & 58.91 & 63.05\\
\hline
Training Step & w/o intra-domain contrastive loss & 99.70 & 82.91 & 60.08 & 62.51\\
Training Step & w/o margin loss & 99.70 & 87.91 & 72.86 & 68.50\\
Training Step & w/o cosine-normalized cross-entropy loss & 99.77 & 77.90 & 69.58 & 72.20 \\

\hline
None &\ours (ours) & 99.70 & 87.14 & 74.74 & 79.10\\

\hline
\end{tabular}
\end{center}

\label{tab:ab_results}
\end{table*}

\begin{figure*}[ht]
	\centering
 \subfigure
        {
        \includegraphics[width=1.7\columnwidth]{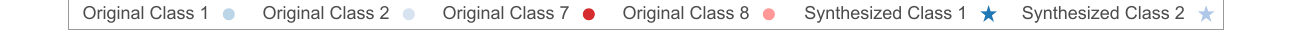}
        \label{fig:legend}
        }
        \hspace{0pt}
\subfigure[w/o synthesis]
        {
        \includegraphics[width=0.55\columnwidth]{CCSI_no_synthesis.pdf}
        \label{fig:no_synth}
        }
        \hspace{15pt}
        \subfigure[w/o mean initialization]
        {
        \includegraphics[width=0.55\columnwidth]{CCSI_no_mean.pdf}
        \label{fig:no_mean}
        }
        \hspace{15pt}
\subfigure[w/o regularization]
        {
        \includegraphics[width=0.55\columnwidth]{CCSI_no_reg.pdf}
        \label{fig:no_reg}
        }
        \hspace{15pt}
        \subfigure[BN]
        {
        \includegraphics[width=0.55\columnwidth]{CCSI_no_CN.pdf}
        \label{fig:no_cn}
        }
        \hspace{15pt}
        \subfigure[\ours (ours)]
        {
        \includegraphics[width=0.55\columnwidth]{CCSI.pdf}
        \label{fig:umap_ours}
        }

	\caption{Visual representations of both original and synthesized images for the BloodMNIST dataset in the latent space, utilizing various methods for the synthesis step using UMAP~\citep{mcinnes2018umap}. We aim to show that the synthetic data generated via \ours{} has the closest distribution with the original ones compared with alternatives. We present the latent space representation of original samples from \emph{previous classes:} the initial two classes (1 and 2), and \emph{current classes:} the two most recently added classes (7 and 8) in the final task, using circle markers ($\bullet$). In addition, we showcase synthesized samples for the initial two classes, represented by star markers ($\star$), which serve as exemplars of original samples that are no longer available.}
	\label{fig:umap}
\end{figure*}

\noindent\textbf{CLBM}~\citep{yang2021continual} uses Gaussian mixture models (GMM) to obtain prototypes for each class and preserve their information. It requires access to a massive prior dataset of the same modality to pre-train the network's feature extractor in the first task before performing continual learning. Our experiment is only able to access such prior data for the Heart Echo dataset. However, CLBM overfits new data and fails to assign distinguishable distributions to them with a final accuracy of 20.2\% in five-way classification.

\noindent\textcolor{black}{\textbf{GPM}~\citep{saha2021gradient} learns new tasks by taking gradient steps in the orthogonal direction to the gradient projection memory, which is computed by analyzing network representations after learning each task using Singular Value Decomposition (SVD). Such an orthogonal update preserves previous data memory even if we do not save the real samples and merely overcome task recency bias. However, it does not guarantee inter-task separation. While we outperform this method on four datasets with a gap of approximately 40\%, it performs better only on the TissueMNIST dataset.} \textcolor{black}{In contrast to GPM, the best-performing methods on other datasets—such as \ours, ABD, LUCIR, and LwF—all use distillation loss to retain knowledge of previous tasks. Therefore, this performance drop can be attributed to the lack of past information retention using gradient projection to save memory.}

\noindent\textbf{ABD}~\citep{smith2021always} uses deep inversion to train a generative model in each task and generate synthetic images for previous task classes. However, it only matches the saved BN means of the trained model with the running BN means of the synthesized data, failing to capture continually introduced information for each class. Additionally, they do not utilize the mean of each class in their generation paradigm, resulting in generated data that is ineffective in representing previous tasks. As a result, ABD underperforms significantly compared to \ours across all datasets, with the most severe gap of 45.9\% and 50.8\% accuracy underperforming over the BloodMNIST and Heart Echo, which have more structural anatomy compared to other datasets. However, we only outperform ABD by 0.7\% over TissueMNIST, which has more diverse samples.

\noindent\textcolor{black}{\textbf{GEM \& AGEM} ~\citep{lopez2017gradient, chaudhry2018efficient} both aim to save a fixed-size memory of previous data for use in further training. However, GEM utilizes a large episodic memory, which is computationally and memory-wise expensive. To address this, AGEM proposes to approximate this process by averaging the saved memory and using a smaller memory, resulting in computational and memory efficiency comparable to EWC.  While this approximation may harm performance compared to GEM, it outperforms oEWC in the BloodMNIST and OrganaMNIST datasets. That being said, despite being data-free, \ours outperforms both GEM and AGEM on all datasets, with a minimum gap of 7\% on TissueMNIST and a maximum gap of 35\% on OrganaMNIST.}

\noindent\textbf{iCarL}~\citep{rebuffi2017icarl} uses a fixed-size memory to store samples from previous tasks. It aims to alleviate forgetting by reminding the network of these stored samples. Although iCarL uses saved memory, we outperform it on BloodMNIST, OraganaMNIST, and Heart Echo datasets by only storing the mean representation of each class with a gap of 11.4\%, 5.6\%, and 33\%, respectively. As shown in Fig.~\ref{fig:SOTApath} and Fig.~\ref{fig:SOTATissue}, our results are comparable to iCarL in PathMNIST and TissueMNIST with a mere decrease of 2.1\% and 4.0\% only after the final task. We attribute this to the low-quality mean images (as shown in the second row of Fig.~\ref{fig:sample_table}) for these datasets, which limits our ability to synthesize high-quality images.

\noindent\textcolor{black}{\textbf{DER \& DER++}~\citep{buzzega2020dark} store samples from previous tasks. However, unlike iCarL, which relies on the network appointed at the end of each task as the sole teaching signal for these samples, DER and DER++ store logits sampled throughout the optimization trajectory. This approach resembles having several different teacher parametrizations. DER++ includes an additional regularization term on top of DER, enhancing the model's performance on previous task samples. While we outperform DER++ on the BloodMNIST and TissueMNIST datasets with a gap of 3\% and 10\%, respectively, it performs slightly better on the PathMNIST and OrganaMNIST datasets, positioning it as the second-best performing non-data-free state-of-the-art model.}

\noindent\textbf{LUCIR}~\citep{hou2019learning} also saves a corset of data points from each seen class, similar to iCarL, but employs bias-correction methods to replay them as anchors of their respective class distribution to maximize the distance between class distributions in the latent space. We evaluate LUCIR in three different scenarios: 1) data-free, 2) non-weighted saved memory, and 3) weighted saved memory. In a data-free scenario, \ours outperforms LUCIR on all datasets except TissueMNSIT, with a maximum gap of 61.7\% in accuracy after the final task over the Heart Echo dataset, demonstrating the effectiveness of our synthesized images. \textcolor{black}{It is also worth mentioning that except for LUCIR, all other methods struggle to perform well on TissueMNIST. Despite LUCIR's good performance on TissueMNIST, it still ranks the lowest among all other datasets, highlighting the complexity of classification in this dataset. This shows that the TissueMNIST dataset poses exceptional challenges, particularly due to the importance of intra-task separation.} When a fixed number of data points are saved as a core set from each class, LUCIR outperforms \ours only on PathMNIST and TissueMNIST with a gap of 5.2\% and 4.2\%, respectively, for the same reasons as explained for iCarL. However, when a weighted sampler is used in LUCIR to increase the impact of the saved samples, it surpasses \ours on all datasets, but with only a gap of 3.6\% and 9.5\% over the final task of BloodMNIST and Heart Echo, respectively, due to inherent information loss in the data-free setting. 

In conclusion, our comparisons demonstrate that our proposed framework, \ours, is highly effective in data-free class incremental learning for medical images and is comparable to state-of-the-art non-data-free methods. It consistently outperforms previous data-free class incremental learning methods, as evidenced by its significantly improved accuracy on all seen classes in the final task on all benchmark MedMNIST datasets and the private Heart Echo dataset, with improvements of up to 51\% in accuracy.

\subsection{Ablation Studies}

To analyze the effect of various components of \ours on the final performance, we conduct an ablation study on BloodMNIST, the dataset with the highest performance among all five datasets. Our study is divided into two parts, and the results are demonstrated in Table~\ref{tab:ab_results}.

\subsubsection{Synthesis Step} \textcolor{black}{Our synthesis step consists of a classification loss for training on real samples and three regularization losses to ensure the quality of generated samples, as stated in Eq.~\ref{eq:reg}. The classification loss ensures the generation of class-specific samples using model inversion, while $\mR_{\rm CN}$  encourages the generation of samples that closely resemble the original data distribution. The  $\mR_{\rm TV}$  and $\mR_{\ell_2}$  regularization terms help reduce noise and pixel-wise variation, as we initialize the pixel values from the noisy mean image of each class.} We try different alternatives for the synthesis step to confirm the synthesized images' impact on the information transformation from previous classes. 
Our goal for data synthesis is not to create samples that look like real data (as shown in Fig.~\ref{fig:sample_table}) but to cover the distribution of previous classes and use them as anchor points to prevent catastrophic forgetting. So, we assess the quality of the \ours samples by mapping the multi-dimensional embedding space of our data to 2D using UMAP~\citep{mcinnes2018umap} (as shown in Fig.~\ref{fig:umap}). UMAP is a dimension-reduction strategy for manifold learning. We first train it using only training data and then use the trained values to transform the embedding space of synthesized data. 

\noindent\textbf{Impact of synthesizing}: We omit the entire synthesis procedure and use only the mean image of classes plus random Gaussian noise to get more various samples as a representative of each class. As shown in Fig.~\ref{fig:no_synth}, mean samples provide a fine anchor point for each class due to the similar anatomical structure of medical data. However, they can not cover the distribution of each class well. So, we encounter a significant drop of 10.5\% compared to \ours. It is worth noting that we are still able to outperform the state-of-the-art providing only the mean of each class, which proves the effectivenesses of losses introduced in the training step to outperform other methods.

\noindent\textbf{Impact of using continual normalization}: 
The results indicate that without the assistance of CN moments, the synthesized images are not qualified well, leading to the largest decrease in final accuracy, with a decrease of 41.2\%. Therefore, distilling knowledge from previously trained networks not only helps improve the initial mean images, but insufficient knowledge during the synthesizing procedure leads to confusion for the network in further steps. Also, the qualitative results in Fig.~\ref{fig:no_reg} show that with matching normalization statics as regularization, generated images are more diverse and can be a better representative of the respective class. Next, we use BN (a commonly used layer in deep networks) instead of CN in model architecture. As BN  is not a class-adaptive normalization, its moments used to synthesize images are recalculated every step only based on new tasks, which leads to training a biased feature extractor and an accuracy drop of 6.2\%. Fig.~\ref{fig:no_cn} also shows that generated samples are closer and more biased toward new classes compared to \ours.

\noindent\textbf{Impact of mean initialization:} The initialization of input images with random Gaussian noise instead of the mean of each class in the synthesis step leads to a 16.0\% drop in the final task, according to quantitative results. Qualitative results presented in Fig.~\ref{fig:no_mean} support this finding, as the synthesized samples are diverse but fail to capture the actual distribution of each class without a good optimization starting point.

\subsubsection{Training Step} As depicted in Fig.~\ref{fig:umap_ours}, the synthesized data offers valuable insights into the prior classes, but they are not equivalent in quality to real data.  To address this issue, we introduce novel losses. We can assess the effects of these losses on training by excluding each one individually.

\noindent\textbf{Impact of domain adaption:} Excluding the intra-domain contrastive loss~(Eq.~\eqref{eq:cosine}) results in a significant decrease of 16.6\% in performance. This can be attributed to the domain discrepancy between the synthesized and original images, which leads to the harmful effects of training over data from two domains.

\noindent\textbf{Impact of intra-task separation:} If we exclude the margin loss~(Eq.~\eqref{eq:margin}), the model will not aim to maximize the distance between synthesized samples from prior tasks and those from new tasks. As a result, there will be less distinct separation among classes from different tasks, which will ultimately result in a reduced 10.6\% accuracy.

\noindent\textbf{Impact of cosine-normalized classifier}: 
We replace the cosine-normalized classification layer with the softmax layer and utilize regular cross-entropy loss to classify the data. However, as the new classes are of superior quality and quantity, the performance of the classifier suffers a substantial drop by 6.9\% due to biased training toward the new data.

To summarize, Table~\ref{tab:ab_results} indicates that the omission of regularization during data synthesis results in the most significant accuracy drop of 41.2\%. This is because optimization regularization plays a crucial role in preventing the divergence of previous class information during continual training. In contrast, the intra-domain contrastive loss is the most effective loss in our pipeline, as removing it leads to a decrease of 16.6\% in the final performance.

\section{Discussion}

To the best of our knowledge, our work \ours{} is the first that leverages model-inversion-based medical image synthesis for data-free class incremental learning in medical imaging. Our contributions not only offer a viable pipeline but also introduce technical novelty in data restoration with appropriate normalization layers and continual learning using novel losses. As summarized in Table~\ref{tab:comp_algo} and evidenced in the experimental results, \ours{} presents several advantages over alternative solutions. 

Particularly, unlike other data synthesis models that utilize a separate generative network to obtain synthetic images via training from accessible data in the same distribution, our approach utilizes the trained classification network and synthesizes images via solving an inverse problem from model weights to pixel space, thus achieving data-free w.r.t. the previous data used to train the model  This makes it more data efficient and avoids common issues such as mode collapse that are often encountered with GAN-based data generation methods.
However, as the model-inversion-based image synthesis strategy directly optimizes the pixel domain, the complexity increases if the image resolution increases due to the challenges with high-dimensional optimization. \textcolor{black}{In this study, we propose initializing the data synthesis process with the mean image of each class as meta-data to improve the fidelity of the generated data. As demonstrated in Fig.~\ref{fig:umap} and Table~\ref{tab:ab_results}, such initialization with mean images significantly enhances performance. Importantly, this initialization does not compromise the data-free nature of our method. Storing one mean image per class does not pose a significant memory storage issue. Additionally, despite being a non-data-free method, using mean data does not violate privacy regulations, as mean images are visually distinct from patients' personal information, as illustrated in Fig.~\ref{fig:sample_table}. Finally, even in cases where the mean class image is not accessible, our ablation studies in Table 5 still demonstrate satisfactory results and outperform data-free methods on the BloodMNIST dataset without mean image initialization.}

Note that our goal is not to create samples that closely resemble real data but to cover previous class distributions and use them as anchor points to prevent catastrophic forgetting. \noindent\textcolor{black}{Since \ours{} is generated by optimizing the main classification (CN-CE) loss used to train the network (Eq.~\eqref{eq:CIloss}), it effectively represents the latent features of real data in the frozen trained network's latent space. Furthermore, we introduced a regularization loss on continual-normalization layers (Eq.~\eqref{eq:r_cn}), which encourages images to remain close to real images not only in the latent space but also throughout the mid-embedding space of the entire classification network. This approach offers a safer option in terms of privacy, as the semantic information in the input space is less susceptible to potential leakage of private patient information compared to other generative models like GANs, which are known to face privacy issues~\citep{karras2020analyzing}.} However, we believe that although our method successfully synthesizes individual samples from their respective distributions, the overall distribution of the generated data is still not diverse enough, as illustrated in Fig.~\ref{fig:umap}. This results in information loss and decreased performance compared to when all the data is fed to the network simultaneously for offline training. We consider this issue as a future direction for investigation, aiming to achieve more diverse synthesized representations of previous classes.
One possible future direction is to incorporate the variance of each class in addition to their mean to generate synthesized samples that cover the entire distribution.
\section{Conclusion}

In this work, we propose \ours, a novel data-free class incremental learning framework for medical image classification. In \ours, we synthesize class-specific images by inverting from the trained model with class-mean image initialization. We explore a recently introduced normalization layer -- CN, to reduce overwriting moments during continual training and propose a novel statistic regularization using the frozen CN moments for image synthesis.
Subsequently, we continue training on new classes and synthesized images using the proposed novel losses to increase the utility of synthesized data by mitigating domain shift between new synthesized and original images of old classes and alleviating catastrophic forgetting and imbalanced data issues among new and past classes. 
Experimental results for four MedMNIST datasets as benchmark public datasets and in-house echocardiography cines as the large-scale and more complex dataset validate that \ours outperforms the state-of-the-art methods in data-free class incremental learning with an improbable gap of up to 51\% accuracy in the final task and get comparable results with the state-of-the-art data-saving rehearsal-based methods.
Our proposed method shows the potential to apply incremental learning in many healthcare applications that cannot save data due to memory constraints or private issues.

\section*{Acknowledgments}
This work is supported by the Natural Sciences and Engineering Research Council of Canada (NSERC), the Canadian Institute of Health Research (CIHR), and Canada CIFAR AI Chair Award.

\bibliographystyle{model2-names.bst}\biboptions{authoryear}
\bibliography{reference.bib}


\end{document}